\documentclass[preprint,5p,compress]{elsarticle}

\usepackage{lscape,caption}
\usepackage{amssymb}
\usepackage{amsmath}
\usepackage{mathtools, cuted}
\usepackage{graphicx, subfigure}
\usepackage{url}
\usepackage{algorithm}
\usepackage{algpseudocode}
\usepackage{color}
\usepackage{booktabs}
\usepackage{stmaryrd}
\usepackage{epstopdf}
\usepackage{pdflscape}
\usepackage{stmaryrd}
\usepackage{color}

\journal{Neurocomputing}

\definecolor{mycolor}{rgb}{0,0,0} % Change to (0,0,0) to remove highlighting of updated text

\begin{document}

\begin{frontmatter}

\title{Tackling Multilabel Imbalance through Label Decoupling and Data Resampling Hybridization}

\author[UGR]{Francisco Charte\corref{cor1}}
\ead{francisco@fcharte.com}

\author[UJA]{Antonio J. Rivera}
\ead{arivera@ujaen.es}

\author[UJA]{Mar\'ia J. del Jesus}
\ead{mjjesus@ujaen.es}

\author[UGR,IND]{Francisco Herrera}
\ead{herrera@ugr.es}

\cortext[cor1]{Corresponding author. Tel.: +34 953 212 892; fax: +34 953 212 472. \\ %Arxiv
	Manuscript accepted at Neurocomputing: \url{https://doi.org/10.1016/j.neucom.2017.01.118} \\
	\textcopyright~2018. This manuscript version is made available under CC  	BY-NC-ND 4.0 license \url{https://creativecommons.org/licenses/by-nc-nd/4.0/}}
\address[UGR]{Department of Computer Science and A.I., University of Granada, 18071 Granada, Spain}
\address[UJA]{Department of Computer Science, University of Ja\'en, 23071 Ja\'en, Spain}
\address[IND]{Faculty of Computing and Information Technology - North Jeddah, King Abdulaziz University, 21589, Jeddah, Saudi Arabia}

\begin{abstract}
The learning from imbalanced data is a deeply studied problem in standard classification and, in recent times, also in multilabel classification. A handful of multilabel resampling methods have been proposed in late years, aiming to balance the labels distribution. However these methods have to face a new obstacle, specific for multilabel data, as is the joint appearance of minority and majority labels in the same data patterns. We proposed recently a new algorithm designed to decouple imbalanced labels concurring in the same instance, called REMEDIAL (\textit{REsampling MultilabEl datasets by Decoupling highly ImbAlanced Labels}). The goal of this work is to propose a procedure to hybridize this method with some of the best resampling algorithms available in the literature, including random oversampling, heuristic undersampling and synthetic sample generation techniques. These hybrid methods are then empirically analyzed, determining how their behavior is influenced by the label decoupling process. As a result, a noteworthy set of guidelines on the combined use of these techniques can be drawn from the conducted experimentation.
\end{abstract}


\end{frontmatter}

\section{Introduction}
Classification \cite{Aggarwal:2014} is one of the most profoundly studied problems in Data Mining, the latter being part of the process known as Knowledge Discovery in Databases \cite{fayyad1996data}. Standard classification tasks comprehend mostly binary and multiclass cases. The main goal is to train, through machine learning algorithms, a model able to automatically classify new incoming data patterns.

Unlike standard classification methods, which produce as output a class label only, multilabel classifiers (MLC) \cite{Tsoumakas3,Charte:SB-MLC,ReviewVentura,TutorialVentura} have to provide a set of relevant labels for each processed instance. MLC has been applied to disease diagnosis in children \cite{Glinka2016}, suggestion of tags for new posts in question answering forums \cite{QUINTA}, image classification \cite{Wei2013}, and identification of multi-functional enzyme \cite{Che2016}, among other tasks. The amount of MLC algorithms proposed in the last decade is impressive.

Imbalanced learning \cite{Alberto:2013,He2009,Lopez:2013,Prati2014} is a well-known problem in binary and multiclass classification, and it also affects multilabel datasets (MLDs). The class frequencies in an imbalanced dataset present large inequalities, a fact that makes harder the learning of an effective classification model. As stated in \cite{Charte:Neucom13}, most MLDs show significant imbalance levels. One of the strategies to face this problem consist in balancing the labels distribution, usually by means of some kind of resampling procedure \cite{JSSanchez:13}. 

The resampling approach is also among the most popular options when it comes to face imbalance learning in the multilabel field. Several resampling algorithms for MLC \cite{Charte:Neucom13,Giraldo:2013,Charte:IDEAL14,Charte:MLSMOTE} has been already proposed in late years, including random oversampling and oversampling, heuristic undersampling and synthetic instance generation solutions.

Multilabel resampling methods have to deal also with some imbalance related specificities of MLDs. One of such problems is described in \cite{Charte:HAIS14} as the concurrence of frequent and rare labels in the same instance. Due to this matter, balancing the labels distribution through resampling techniques becomes harder since removing instances with majority labels will also imply the loss of minority ones. Analogously, adding new instances by cloning existing ones would increase the frequency of already common labels. 

In \cite{Charte:REMEDIAL} we proposed a specialized method to solve this problem, the REMEDIAL (\textit{REsampling MultilabEl datasets by Decoupling highly ImbAlanced Labels}) algorithm. It works by decoupling imbalanced labels, as will be further detailed, and its effectiveness in its own, as long as it is applied to MLDs having a certain concurrence level, has been already proven. Considering that REMEDIAL separates majority and minority labels in independent samples, it is reasonable to expect that resampling methods would be able to better balance the labels distribution.

Our starting hypothesis is that REMEDIAL can improve the behavior of standard multilabel resampling methods, performing a prior label decoupling where needed. Founded on this hypothesis, the goal in this work is to hybridize REMEDIAL with some of the resampling methods already described in the literature, aiming to improve current imbalanced multilabel learning results. Our premise is that oversampling algorithms would be able to perform a better work, generating more new instances from existing ones having only minority labels, instead of a mixture of minority and majority ones. In the same way, undersampling methods should be capable of removing instances containing only majority labels, avoiding the information loss which implies deleting samples with label concurrence.

Specifically, we propose three hybrid methods based on as many resampling algorithms, named ML-ROS \cite{Charte:Neucom13}, MLeNN \cite{Charte:IDEAL14} and MLSMOTE \cite{Charte:MLSMOTE}. All of them will be further detailed, and their behavior will be tested through an extensive experimentation, including ten popular MLDs, three disparate MLC algorithms and five distinct evaluation metrics. The purpose of the study can be summarized into two main objectives:

\begin{itemize}
    \item First, determine if the hybridization of REMEDIAL with a resampling algorithm could produce an improvement in classification results. For doing so, three MLCs will be used to process a common set of ten MLDs in two versions, one preprocessed with the resampling methods on their own and another one also with their respective hybrid versions.
    
    \item Second, analyze the potential interactions between the resampling methods, the MLCs and the MLDs' traits. This way, a group of rules about when these hybrid versions would be useful could be drawn.
\end{itemize} 

The remainder of this paper is organized into the following sections. In Section \ref{Background} the multilabel classification task is introduced, along with the specific details related to imbalanced learning in this field. In Section \ref{Proposal} the proposed hybridization is described. The conducted experimentation, results and analysis are provided in Section \ref{Experimentation}. Lastly, Section \ref{Conclusions} state the final conclusions.

\section{Background}\label{Background}
In this section the multilabel classification task is briefly introduced and put into context. The obstacles while dealing with imbalanced MLDs, as well the ways they have been faced, are also described.

Multilabel learning \cite{Tsoumakas3,Charte:SB-MLC,ReviewVentura,TutorialVentura} is currently a very active field. The techniques for multilabel classification have been applied to text categorization \cite{Elghazel2016}, image annotation \cite{Jing2016}, tag suggestion \cite{QUINTA} for question answering forums, and disease diagnosis in children \cite{Glinka2016}, among others tasks. All these problems have a common characteristic, each one of the data patterns is linked to several labels at once, instead of only one class as in standard classification. There is a global set of labels $\mathcal{L}$, containing all $k$ labels used in the $\mathcal{D}$. $\mathcal{D}$ being an MLD, $\mathcal{D}_i$ would be its \textit{i-th} instance, and $Y_i \subseteq \mathcal{L}$ the subset of labels (labelset) which are relevant to that instance. The role of any MLC is to provide $Z_i \subseteq \mathcal{L}$ with the labels predicted for new instances, with goal of being as close as possible to $Z_i = Y_i$. 

Since each instance in an MLD is associated only to a subset of $\mathcal{L}$, some metrics have been defined to assess the degree of multilabelness of MLDs. The most common ones are label cardinality, i.e. $Card$ (\ref{Card}), and label density, i.e. $Dens$ (\ref{Dens}).

    \begin{equation} 
        Card\left(D\right) = \frac{1}{n} \displaystyle\sum\limits_{i=1}^{n} \lvert Y_i\rvert 
        \label{Card}
    \end{equation}

    \begin{equation} 
        Dens\left(D\right) = \frac{1}{k} \frac{1}{n} \displaystyle\sum\limits_{i=1}^{n} \lvert Y_i\rvert 
        \label{Dens}
    \end{equation}

\subsection{MLC Approaches}
Classification of multilabel data is usually tackled by means of data transformation or method adaptation techniques. These are the two common approaches followed by most proposals. The former aims to transform the original multilabel task into one or more standard classification tasks, while the latter intents to adapt the standard classification models to make them able to work natively with multilabel data.

The two best known data transformation methods are BR (\emph{Binary Relevance}) \cite{Godbole} and LP (\emph{Label Powerset}) \cite{Boutell}. BR produces a set of binary datasets from the original MLD. Then, each binary dataset is processed by a standard classifier. Eventually, the individual predictions are merged \cite{mencia2010efficient} to obtain the subset of labels relevant to each test instance. LP takes each possible label combination as class identifier, transforming the original MLD into a multiclass dataset. After using it to train a standard classifier, the predicted classes are back transformed to subsets of labels. Both BR and LP are the foundation for many multilabel ensemble-based methods.

Regarding the second mentioned approach, MLC algorithms based on many of the standard classification methods have been proposed in the literature. Among them, there are MLC adaptations of the C4.5 tree induction algorithm \cite{Clare}, instance-based classifiers such as ML-kNN \cite{Zhang1}, SVM adaptations as the one proposed in \cite{Elisseeff1}, multilabel neural networks \cite{Zhou:MIML:2009}, etc. An extensive review on multilabel classification techniques is provided in \cite{ReviewVentura}.

\subsection{Learning from Imbalanced MLDs}
The learning from imbalanced data is deeply studied problem in standard classification. As stated in \cite{Lopez:2013}, it has been mainly confronted trough data resampling, classifier adaptation and cost-sensitive techniques. When it comes to classify imbalanced MLDs, aside from data resampling \cite{Charte:Neucom13,Giraldo:2013,Charte:IDEAL14,Charte:MLSMOTE,Tahir:2012} and classifier adaptation \cite{Tepvorachai:2008,He:2012,LISHI:2013} the ensemble-based approach has been also explored \cite{Tahir:2012:2}. 

An imbalanced MLD presents large differences among the labels distributions, so that some of them are very frequent (majority labels) while other ones are quite rare (minority ones). To assess these differences the \emph{IRLbl} (\ref{IRLbl}) and \emph{MeanIR} (\ref{MeanIR}) metrics were proposed in \cite{Charte:Neucom13}. The symbol $\llbracket \rrbracket$ denotes de Iverson bracket, which returns 1 if the expression inside it is true or 0 otherwise. The \emph{IRLbl} is evaluated for each label in $\mathcal{L}$, and provides an individual imbalance level. The global imbalance or \emph{MeanIR} is obtained by averaging the \emph{IRLbl} for all labels.

    \begin{equation} \textit{IRLbl(l)} = 
   \frac{
      \displaystyle\max\limits_{l' \in \mathcal{L}}^{}
      	\left(\displaystyle\sum\limits_{i=1}^{|D|}{\llbracket l' \in Y_i \rrbracket}\right)
   }
   {
      \displaystyle\sum\limits_{i=1}^{|D|}{\llbracket l \in Y_i \rrbracket}} . 
   \label{IRLbl}
   \end{equation}

    \begin{equation}
	\textit{MeanIR} = \frac{1}{|\mathcal{L}|} \displaystyle\sum\limits_{l \in \mathcal{L}}^{}\textit{IRLbl(l)} .
    \label{MeanIR}
    \end{equation}
  
Some of the resampling methods adapted to deal with multilabel data have been random undersampling and oversampling, heuristic undersampling, and synthetic instance generation. Several of these proposals where recently compared in \cite{Charte:MLSMOTE}. The following algorithms are mong the best performers:
\begin{itemize}
    \item \textbf{ML-ROS:} It was introduced in \cite{Charte:Neucom13} as a way to balance label distribution through random oversampling. As it can be observed in Alg. \ref{ML-ROS}, it takes into account the presence of several minority labels, randomly looks for instances associated to them and generates clones of these instances. As can be seen in Alg. \ref{ML-ROS}, the amount of clones created by the method is set as a percentage relative to the total number of samples in the MLD.
    
\begin{algorithm}[ht!]
\small
\caption{ML-ROS algorithm's pseudo-code.}
\label{ML-ROS}
\begin{algorithmic}[1]
\algnewcommand{\LineComment}[1]{\State \(\triangleright\) #1}

\Statex \textbf{Inputs}: $<$Dataset$>$ $D$, $<$Percentage$>$ $P$
\Statex \textbf{Outputs}: Preprocessed dataset
\Statex
\State $samplesToClone\gets |D| / 100 * P$\Comment{P\% size increment}
\State $L \gets$ labelsInDataset($D$) \Comment{Obtain the full set of labels}
\State \textit{MeanIR} $\gets$ calculateMeanIR($D, L$)
\For{\textbf{each} $label$ \textbf{in} $L$} \Comment{Bags of minority labels samples}
    \State \textit{IRLbl}$_{label} \gets$ calculateIRperLabel($D, label$)
	\If{\textit{IRLbl}$_{label} >$ \textit{MeanIR}}
	  \State $minBag_{i++} \gets Bag_{label}$
	\EndIf
\EndFor

\While{$samplesToClone > 0$} \Comment{Instances cloning loop}
   \LineComment{Clone a random sample from each minority bag}
   \For{\textbf{each} $minBag_i$ \textbf{in} $minBag$} 
    \State $x \gets$ random($1,|minBag_i|$)
    \State $D \gets D +$ cloneSample($minBag_i, x$)
    \If{\textit{IRLbl}$_{minBag_i} <= $\textit{MeanIR}}
    	\State $minBag \rightarrow minBag_i$ \Comment{Exclude from cloning}
    \EndIf
    \State - -$samplesToClone$
   \EndFor
\EndWhile
\State \textbf{return} $D$
\normalsize
\end{algorithmic}
\end{algorithm}

    \item \textbf{MLeNN:} Presented in \cite{Charte:IDEAL14}, it is an undersampling algorithm based on the well-known ENN (\emph{Edited Nearest Neighbor}) rule. Its pseudo-code is provided in Alg. \ref{MLeNN}. Those samples containing only majority labels and whose labelset is in discordance with that of their neighbors are removed. 
    
\begin{algorithm}[ht!]
\small
\caption{MLeNN algorithm pseudo-code.}
\label{MLeNN}
\begin{algorithmic}[1]
\algnewcommand{\LineComment}[1]{\State \(\triangleright\) #1}

\Statex \textbf{Inputs}: $<$Dataset$>$ $D$, $<$Threshold$>$ \textit{HT}, 
\Statex \hspace{4em}$<$NumNeighbors$>$ \textit{NN}
\Statex \textbf{Outputs}: Preprocessed dataset
\Statex
\For{\textbf{each} $sample$ \textbf{in} $D$}
  \For{\textbf{each} $label$ \textbf{in} $getLabelset(D)$} 
	\If{IRLbl($label$) $>$ \textit{MeanIR}}
    	\LineComment{Preserve instance with minority labels}
		\State Jump to next sample 
	\EndIf
  \EndFor
  \State \textit{numDifferences} $\gets 0$
  \For{\textbf{each} $neighbor$ \textbf{in} nearestNNs(\textit{sample}, \textit{NN})}
  	\If{HammingDist(\textit{sample}, \textit{neighbor}) $>$ \textit{HT}}
  		\State \textit{numDifferences} $\gets$ \textit{numDifferences}$+1$
  	\EndIf
  \EndFor
  \If{\textit{numDifferences}$\geq$\textit{NN}$/2$}
    \State markForRemoving(\textit{sample})
  \EndIf
\EndFor
\State deleteAllMarkedSamples($D$)
\State \textbf{return} $D$
\normalsize
\end{algorithmic}
\end{algorithm}

    \item \textbf{MLSMOTE:} It was proposed in \cite{Charte:MLSMOTE}. This algorithm is founded on the popular SMOTE (\emph{Synthetic Minority Over-sampling Technique}) algorithm. As ML-ROS it considers several minority labels, instead of only one as the original SMOTE. Once the instances in which these labels appear have been found, new instances are generated with synthetic attributes and also synthetic labelsets, both produced from the information of their nearest neighbors. As can be observed in Alg. \ref{MLSMOTEAlg} (lines 5-7), MLSMOTE only takes as seeds the instances in which some minority label appears. Then, their nearest neighbors are located. One of them is randomly picked to produce the synthetic set of input attributes. Lastly, all of them serve as reference to generate the synthetic labelset. Additional details about MLSMOTE implementation can be found in \cite{Charte:MLSMOTE}.
    
\begin{algorithm}[ht!]
\small
%\footnotesize 
\caption{MLSMOTE algorithm's pseudo-code.}
\label{MLSMOTEAlg}
\algnewcommand{\LineComment}[1]{\State \(\triangleright\) #1}
\begin{algorithmic}[1]
\Statex \textbf{Inputs}: $<$Dataset$>$ $D$, $<$NumNeighbors$>$ \textit{k}
\Statex \textbf{Outputs}: Preprocessed dataset
\Statex
\end{algorithmic}

\begin{algorithmic}[1]
\State $L \gets$ labelsInDataset($D$) \Comment{Full set of labels}
\State \textit{MeanIR} $\gets$ calculateMeanIR($D, L$)
\For{\textbf{each} $label$ \textbf{in} $L$} 
    \State \textit{IRLbl}$_{label} \gets$ calculateIRperLabel($D, label$)
	\If{\textit{IRLbl}$_{label} >$ \textit{MeanIR}}
	  \LineComment{Bags of minority labels samples}
	  \State \textit{minBag} $\gets$ getAllInstancesOfLabel(\textit{label}) 
	  %\State \textit{mapVDM} $\gets$ genMapVDM(\textit{label})
	  \For{\textbf{each} $sample$ \textbf{in} $minBag$} 
	  	\State \textit{distances} $\gets$ calcDistance(\textit{sample}, \textit{minBag}) 
	  	\State sortSmallerToLargest(\textit{distances})
	  	\LineComment{Neighbor set selection}
	  	\State \textit{neighbors} $\gets$ getHeadItems(\textit{distances}, \textit{k}) 
	  	\State \textit{refNeigh} $\gets$ getRandNeighbor(\textit{neighbors})
	  	\LineComment{Feature set and labelset generation}
	  	\State \textit{synthSmpl} $\gets$ newSample(\textit{sample}, 
	  	\State \hspace{7em}\textit{refNeigh}, \textit{neighbors})
	  	\State $D = D$ + \textit{synthSmpl}
	  \EndFor
	\EndIf
\EndFor
\State \textbf{return} $D$

\Statex
\Function{newSample}{\textit{sample}, \textit{refNeigh}, \textit{neighbors}}
	\State \textit{synthSmpl} $\gets$ \textbf{new} Sample \Comment New empty instance
	\LineComment{Feature set assignment}
	\For{\textbf{each} \textit{feat} \textbf{in} $synthSmpl$}
	  \If{typeOf(\textit{feat}) is numeric}
	    \State \textit{diff} $\gets$ \textit{refNeigh.feat} - \textit{sample.feat}
	    \State \textit{offset} $\gets$ \textit{diff} * randInInterval(0,1)
	    \State \textit{value} $\gets$ \textit{sample.feat} + \textit{offset}
	  \Else
	  	\State \textit{value} $\gets$ mostFreqVal(\textit{neighbors},\textit{feat})
	  \EndIf
	  \State \textit{syntSmpl.feat} $\gets$ \textit{value}
	\EndFor
	
	\LineComment{Label set assignment}
		  \State \textit{lblCounts} $\gets$ counts(sample.labels)
		  \State \textit{lblCounts} $+\gets$ counts(neighbors.labels)
		  \State \textit{labels} $\gets$ \textit{lblCounts} $>$ \textit{(k+1)} / 2
	\State \textit{synthSmpl.labels} $\gets$ \textit{labels}
	\State \textbf{return} \textit{synthSmpl}
\EndFunction
\normalsize
\end{algorithmic}
\end{algorithm}
   
\end{itemize}

\subsection{Concurrence Among Imbalanced Labels}
\begin{figure*}[ht!]
\mbox{
\includegraphics[width = \columnwidth]{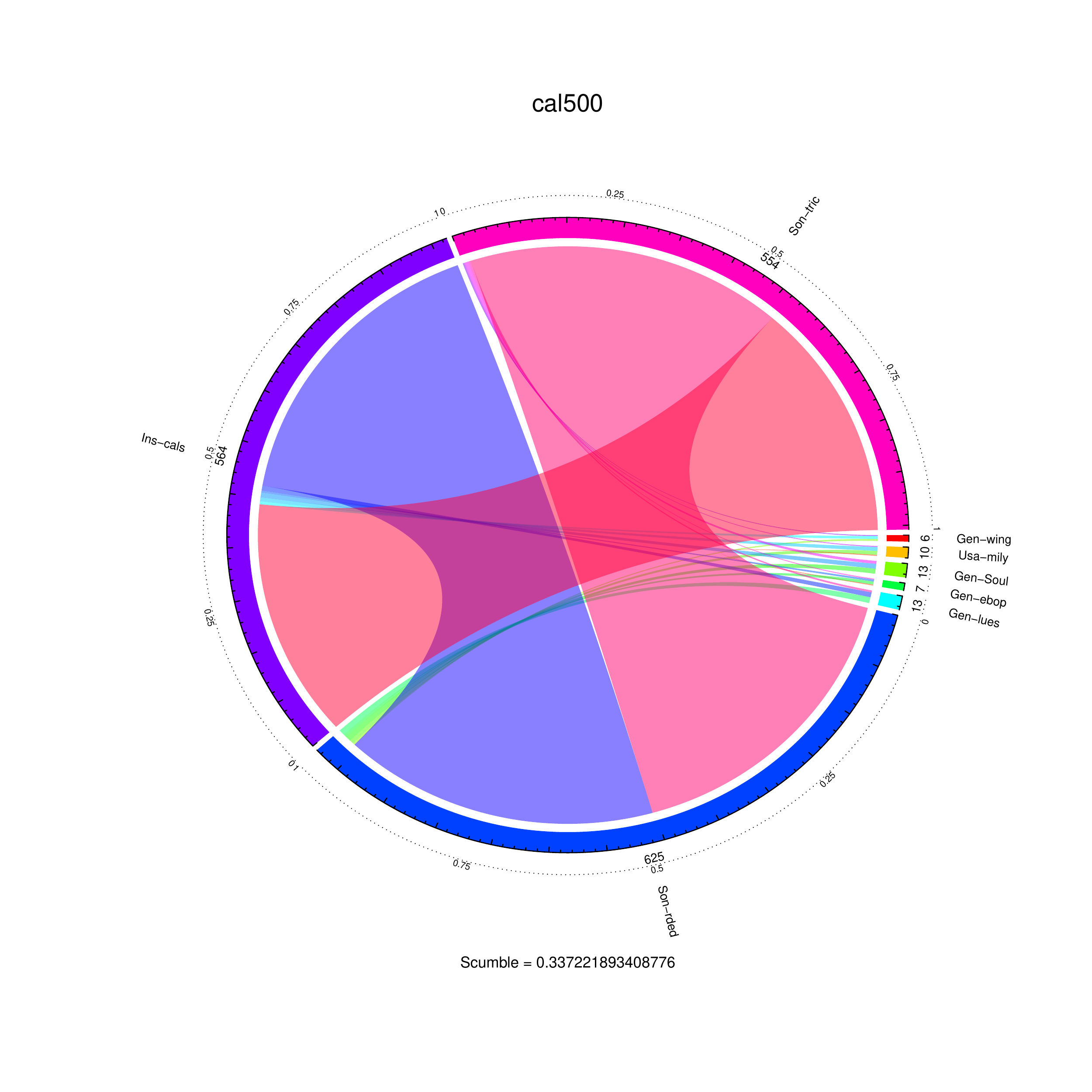}
\includegraphics[width = \columnwidth]{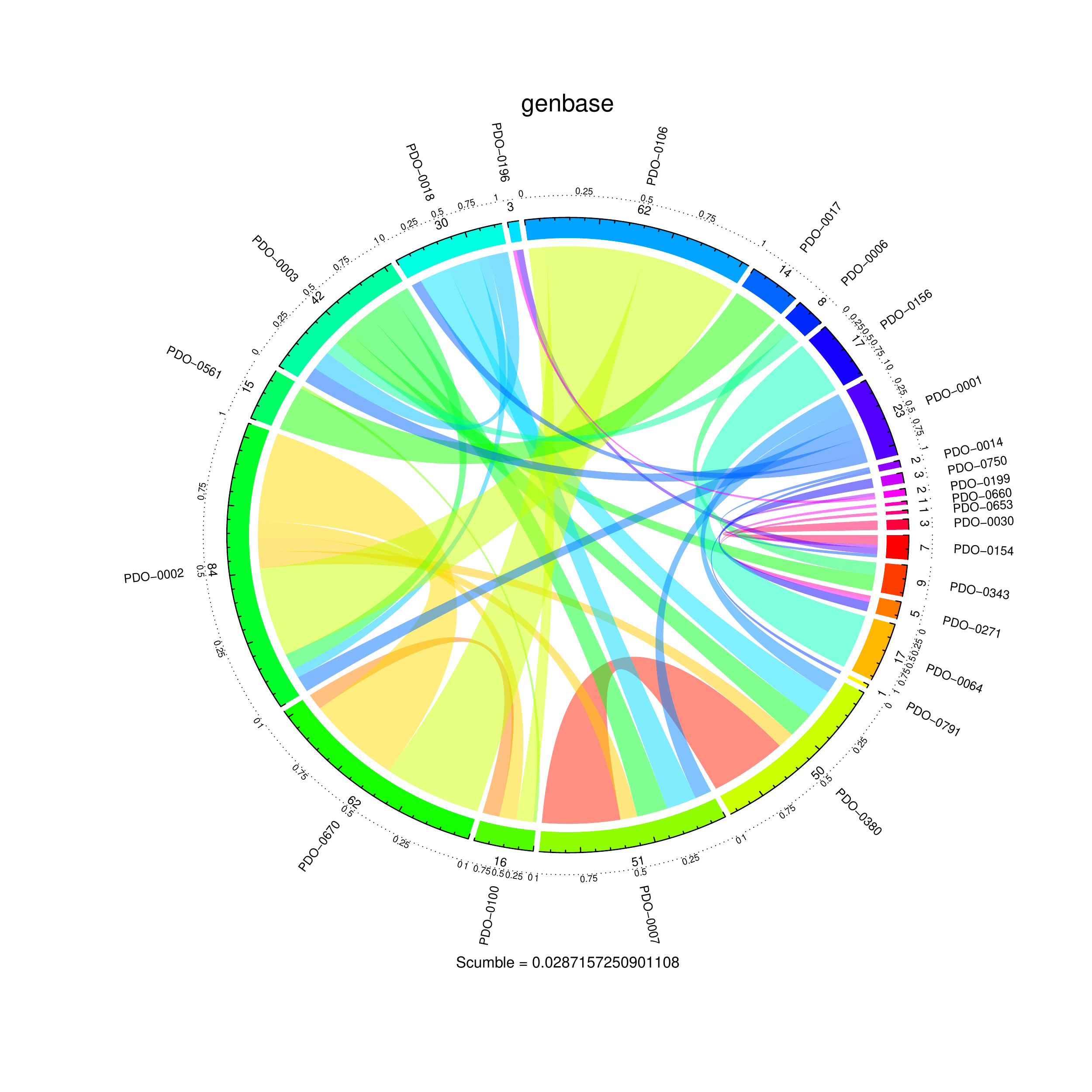}
}
\caption{Concurrence among minority and majority labels in four MLDs.}
\label{Concurrence}
\end{figure*}

Since each instance in an MLD has two or more labels, it is not rare that some of them are very common ones while others are minority labels. This fact can be depicted using an interaction plot\footnote{Visualizing all label interactions in an MLD is, in some cases, almost impossible due to the large number of labels. For that reason, only the most frequent labels and the most rare ones for each MLD are represented in these plots. High resolution versions of these plots can be generated using the \texttt{mldr}  R package \cite{Charte:mldr}.} as the ones shown in Fig. \ref{Concurrence}. In addition, the level of concurrence between common and rare labels can be assessed through the \emph{SCUMBLE} metric \cite{Charte:HAIS14}, defined in (\ref{SCUMBLE}). This metric provides a simple to understand concurrence indicator, whose values will be in the $[0,1]$ range. The higher the value the more instances containing minority and majority labels exist in the MLD.

\begin{equation}
  \textit{SCUMBLE}\left(D\right) = \frac{1}{n}
   \displaystyle\sum\limits_{i=1}^{n} \textit{SCUMBLE}_{ins}\left(i\right)
\label{SCUMBLE}
\end{equation}

\begin{equation}
  \textit{SCUMBLE}_{ins}\left(i\right) =
     1 - \frac{1}{\overline{\textit{IRLbl}_i}}\left(\prod\limits_{l \in L}^{} \textit{IRLbl}_{il}\right)^{\left(1/k\right)}
\label{SCUMBLEIns}
\end{equation}

The left plot of the aforementioned figure corresponds to an MLD with a high level of concurrence between imbalanced labels, denoted by a \textit{SCUMBLE} above of 0.1. As can be seen, the minority labels (on the right side) are entirely linked with some majority labels. In some MLDs the concurrence between majority and minority labels is low, as shown in the right plot of Fig. \ref{Concurrence}. In these cases the level of \textit{SCUMBLE} is below 0.1, and as can be seen there are many arcs between minority labels, denoting interactions between them but not with the majority ones.

The aforementioned multilabel resampling algorithms will not have an easy work while dealing with MLDs which have a high \textit{SCUMBLE}. Undersampling algorithms can produce a loss of essential information, as the samples selected for removal because majority labels appear in them can also contain minority labels. In the same way, oversampling algorithms limited to cloning the labelsets, such as the proposals in \cite{Giraldo:2013,Charte:Neucom13}, can be also increasing the presence of majority labels. These facts were empirically demonstrated in \cite{Charte:HAIS14}.

\begin{algorithm}[ht!] 
\caption{REMEDIAL algorithm's pseudo-code.}
\label{REMEDIALAlg}
\begin{algorithmic}[1]
\algnewcommand{\LineComment}[1]{\State \(\triangleright\) #1}

\Statex \textbf{Inputs}: $<$Dataset$>$ $D$, $<$Labels$>$ $L$
\Statex \textbf{Outputs}: Preprocessed dataset
\Statex
\LineComment{Calculate imbalance levels}
\State \textit{IRLbl$_l$} $\gets$ calculateIRLbl($l$ in $L$) 
\State \textit{IRMean} $\gets {\overline{IRLbl}}$
\LineComment{Calculate SCUMBLE}
\State \textit{SCUMBLEIns$_i$} $\gets$ calculateSCUMBLE($D_i$ in $D$) 
\State \textit{SCUMBLE} $\gets {\overline{\textit{SCUMBLEIns}}}$
\For{\textbf{each} \textit{instance i} \textbf{in} $D$}
     	\If{\textit{SCUMBLEIns$_i$} $> \textit{SCUMBLE}$ }
   	  \State $D'_i \gets D_i$ \Comment{Clone the affected instance}
   	  \LineComment{Maintain minority labels}
   	  \State $D_i[labels_{\textit{IRLbl} <= \textit{IRMean}}] \gets 0$ 
   	  \LineComment{Maintain majority labels}
   	  \State $D'_i[labels_{\textit{IRLbl} > \textit{IRMean}}] \gets 0$ 
   	  \State $D \gets D + D'_i$
     	\EndIf
\EndFor
\State \textbf{return} $D$
\normalsize
\end{algorithmic}
\end{algorithm}

The imbalanced labels concurrence in MLDs can be alleviated through a label decoupling strategy, as described in \cite{Charte:REMEDIAL}. The proposed algorithm is called REMEDIAL and its pseudo-code is shown in Alg. \ref{REMEDIALAlg}. What it does is looking for instances having a high \emph{SCUMBLE} level, i.e. it contains both majority and minority labels. The decoupling of these data samples consist in cloning them, obtaining a couple of instances in which one will be associated to the majority labels and the other one to the minority labels. This way the level of concurrence is reduced. 

\section{Adapting REMEDIAL: Label Decoupling and Data Resampling Hybridization}\label{Proposal}
In this section the procedure to hybridize label decoupling with data resampling methods is presented, and three hybrid versions are detailed. These will be empirically tested in the following section.

\begin{figure*}[ht!]
\includegraphics[width = \textwidth]{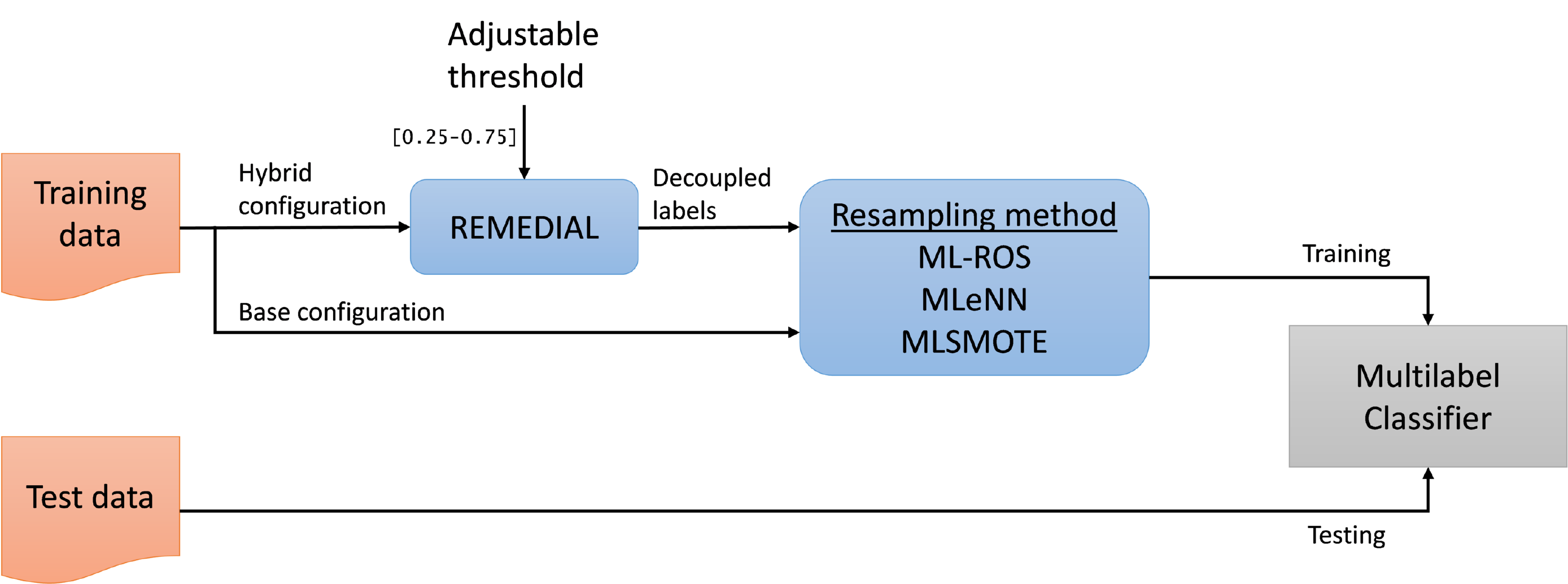}
\caption{Structure of the proposed hybridization.}
\label{Hybridization}
\end{figure*}

As stated in \cite{Charte:REMEDIAL}, REMEDIAL is able to improve classification results on its own, as far as it is applied to MLDs having a high \emph{SCUMBLE} value. In the following, how this technique can be combined with standard resampling methods is analyzed. For doing so, an adapted version of REMEDIAL is going to be used.  In the original version of REMEDIAL a fixed threshold is used to decide which instances are decoupled. This threshold is set to the average \emph{SCUMBLE} value. Our adapted version takes the threshold as another input parameter, so that different cut points can be studied. In particular, the percentiles .25, .37, .50, .62, and .75 will be used, percentile .50 being equivalent to the default threshold in REMEDIAL. The goal is to study if there is a certain optimum threshold for REMEDIAL. The block structure of this hybridization is the one shown in Fig. \ref{Hybridization}.

\subsection{REMEDIAL Hybridizing with Random Oversampling}
As was experimentally demonstrated in \cite{Charte:Neucom13}, random oversampling can deliver improvements in multilabel classification performance. The ML-ROS algorithm in particular is able to balance label distribution following a relatively simple approach. It looks for instances associated to some of the minority labels, then generates clones from these instances.  Although ML-ROS performs well in many scenarios, it would be not able to balance the label distribution of MLDs having a high imbalanced labels concurrence. Due to the joint appearance of minority and majority labels, most of the samples picked by the method would be also associated to frequent labels and, therefore, the clones will include both types of labels.

The proposed hybrid method will firstly decouple imbalanced labels, then will look for instances linked to minority labels and will produce clones from them. These new samples will increase the frequency of rare labels without also implying a grow in those linked to majority labels. As a result, the MLDs would have a more balanced label distribution and would be easier to process by MLC algorithms.

\subsection{REMEDIAL Hybridizing with Heuristic \\Undersampling}
Undersampling methods rely on removing instances in which majority labels appear. It is something that must be done with extreme care. Since each sample in an MLD contains several labels, choosing those having majority labels could also affect the minority ones. For this reason MLeNN \cite{Charte:IDEAL14} starts by excluding all the samples in which some minority labels appear. Then, the remainder instances are processed one by one comparing their labelsets with those of their nearest neighbors. The samples significantly different from more than half of their neighbors are removed from the MLD.

Since MLeNN excludes from processing the instances linked to minority labels, which are potentially also associated to majority ones, there will be a certain amount of samples that never will be evaluated, whether they are similar to their neighbors or not. The hybrid version, by firstly applying the decoupling process, will have more instances acting as candidates to be removed. All the patterns that previously contained minority labels, but now are exclusively associated to majority ones, will be processed by MLeNN instead of being oversighted.

\subsection{REMEDIAL Hybridizing with Synthetic Instance \\Generation}
In \cite{Charte:MLSMOTE} an adaptation of SMOTE to work with multilabel data, called MLSMOTE, was proposed. Unlike SMOTE, MLSMOTE considers the presence of a set of minority labels, instead of only one class. In addition to the synthetic attributes, MLSMOTE also produces a synthetic labelset for the new instances. The set of labels is obtained from a ranking of labels present in the instance being processed and its nearest neighbors. 

The decoupling procedure, applied before the minority samples are selected, will influence the labelset of the chosen sample and potentially also those of its neighbors. As a result, the synthetic instance generated by MLSMOTE will be assigned a slightly different set of labels, sometimes without introducing the majority labels that were present in the sample before decoupling them. The consequence should be a more balanced label distribution, able to induce a better classification model than the base configuration.

\section{Experimentation}\label{Experimentation}
In this section the three hybrid preprocessing methods just described are tested, and they are compared against the original versions of each resampling method. First, the experimental test bed is detailed. Then, the obtained results are provided. Lastly, these results are discussed and analyzed.

\subsection{Experimental Framework}
In order to check how the proposed hybridization influences the behavior of each resampling method ten MLDs from disparate application fields has been used. All of them can be obtained from the R Ultimate Multilabel Repository \cite{Charte:RUMDR}, along with their respective references. Their fundamental traits are provided in Table \ref{MLDs}. The meaning of each column, from left to right, is the following:

\begin{table*}[ht!]
\centering
\setlength\tabcolsep{0.5em}
\begin{tabular}{lrrrrrrrcr}
\toprule
\textbf{Dataset} & \multicolumn{1}{c}{\textbf{Inst.}}  & \multicolumn{1}{c}{\textbf{Attr.}} & \multicolumn{1}{c}{\textbf{Labels}} & \multicolumn{1}{c}{\textbf{LSet}} & \multicolumn{1}{c}{\textbf{Card}} & \multicolumn{1}{c}{\textbf{Dens}} & \multicolumn{1}{c}{\textbf{MeanIR}} & \multicolumn{1}{c}{\textbf{SCUMBLE}} & \multicolumn{1}{c}{\textbf{TCS}} \\ 
\midrule
{yeast}       &  2 417 &    103 &  14 &    198 &  4.237 & 0.303 &   7.197 & 0.104 & 12.562  \\ 
{cal500}      &    502 &     68 & 174 &    502 & 26.044 & 0.150 &  20.578 & 0.337 & 15.597  \\
{medical}     &    978 &  1 449 &  45 &     94 &  1.245 & 0.028 &  89.501 & 0.047 & 15.629  \\
{tmc2007}     & 28 596 & 49 060 &  22 &  1 341 &  2.158 & 0.098 &  15.157 & 0.175 & 16.372  \\
{enron}       &  1 702 &  1 001 &  53 &    753 &  3.378 & 0.064 &  73.953 & 0.303 & 17.503  \\ 
{mediamill}   & 43 907 &    120 & 101 &  6 555 &  4.376 & 0.043 & 256.405 & 0.355 & 18.191  \\
{chess}       &  1 675 &    585 & 227 &  1 078 &  2.411 & 0.011 &  85.790 & 0.262 & 18.779  \\ 
{corel16k}    & 13 766 &    500 & 153 &  4 803 &  2.859 & 0.019 &  34.155 & 0.273 & 19.722  \\ 
{corel5k}     &  5 000 &    499 & 374 &  3 175 &  3.522 & 0.009 & 189.568 & 0.394 & 20.200  \\ 
{delicious}   & 16 105 &    500 & 983 & 15 806 & 19.017 & 0.019 &  71.052 & 0.532 & 22.773  \\
\bottomrule
\end{tabular}
\caption{Main characteristics of the MLDs used in the experimental study.}
\label{MLDs}
\end{table*}

\begin{itemize}
    \item \textbf{Dataset:} The usual name the MLD is known in the literature.
    \item \textbf{Inst.:} Number of data instances in the MLD.
    \item \textbf{Attr.:} Number of input attributes.
    \item \textbf{Labels:} Total number of labels.
    \item \textbf{LSet:} Number of distinct label combinations in the MLD.
    \item \textbf{Card:} Label cardinality.
    \item \textbf{Dens:} Label density.
    \item \textbf{MeanIR:} Average imbalance ratio.
    \item \textbf{SCUMBLE:} Level of imbalanced labels concurrence.
    \item \textbf{TCS:} Theoretical Complexity Score. This metric is defined in \cite{Charte:HAIS16} as a way of assessing the complexity of MLDs.
\end{itemize}

The MLDs are shown in the table ordered according to their TCS value, from the simplest one to the most complex. Theoretically, the higher is the value the harder will be for the classifier to correctly predict labelsets for new instances. These datasets were partitioned into 10 folds, following the stratified partitioning strategy described in \cite{Charte:HAIS16}, and the usual cross validation scheme was used to collect the results. The training partitions of each dataset were preprocessed using six configurations. The first one applies the original resampling method, ML-ROS, MLeNN or MLSMOTE. The other five correspond to the respective hybrid version with the different thresholds for REMEDIAL previously enumerated. So, there will be 18 distinct configurations for each MLD.

Regarding the classifiers the prior MLD configurations are going to be processed with, the goal is to use some of the ones having more influence in the field. Due to this reason, the following three algorithms were chosen:
\begin{itemize}
    \item \textbf{BR:} Binary Relevance \cite{Godbole} is the most popular transformation method for multilabel data. As explained before, it trains an independent binary classifier for each label, then joins the individual predictions to obtain the final labelset. Despite its apparent simplicity, BR is usually among the best performers. Above all, BR is the foundation of many other MLC algorithms, including several ensemble-based solutions such as CC/ECC (\emph{Classifier Chains/Ensemble of Classifier Chains}) \cite{Read}, RPC (\emph{Ranking by Pair-wise Comparison}) \cite{RPC} and CLR (\emph{Calibrated Label Ranking}) \cite{CLR}. Therefore, the analysis of results obtained with BR could be extrapolated to many other MLC proposals to a certain extent.
    
    \item \textbf{LP:}  Label Powerset \cite{Boutell} is also a straightforward way of facing multilabel classification, simply by taking each label combination as an individual class identifier. This approach has been also used as a starting point for some other MLC algorithms, including PS (\emph{Pruned Sets}) \cite{Read:2008}, EPS (\emph{Ensemble of Pruned Sets}) \cite{Read:2008:2}, HOMER (\emph{Hierarchy of Multilabel Classifiers}) \cite{HOMER} and RAkEL (\emph{Random k-Labelsets}) \cite{Tsoumakas4}, among others. Thus, the conclusions drawn from LP could also be applicable to all these proposals.
    
    \item \textbf{ML-kNN:} This instance-based classification algorithm was proposed in \cite{Zhang1}, and it has been the foundation for other more advanced classifiers, such as IBLR-ML \cite{Cheng}. Essentially, ML-kNN computes the a priori probabilities for each label, and uses this information later, when a test sample arrives, to calculate the conditional probabilities and thus obtain the predicted labelset. It is included in the experimentation by the same reasons given above, it is a simple method and have influenced many others.
\end{itemize}

The implementation of this three algorithms can be found in MULAN \cite{MULAN}. This was the software tool used to conduct the described experiments.

The predictions given by the aforementioned three MLC methods are going to be assessed with several multilabel evaluation metrics, since each one of them provides a different perspective of the classifier performance. Hamming Loss (\ref{HL}) and Ranking Loss (\ref{RankingLoss}) are among the most usual multilabel performance metrics, included in most studies. Both are loss metrics, so the goal has to be to minimize them. Precision (\ref{Precision}) and F-measure (\ref{F1}) are common evaluation metrics in classification problems, as it is AUC (\ref{MicroAUC}) (\emph{Area Under the ROC Curve}).  The three of them are performance metrics, so the goal is to maximize them.
\begin{itemize}
    \item \textbf{Hamming Loss:} It iss the most popular multilabel evaluation metric. It counts the number of missclassifications, whether they are false positives or false negatives, and then averages by the amount of instances and labels. 
    
    \item \textbf{Ranking Loss:} It measures the proportion of times in which a non-relevant label is ranked above a true relevant one, so the lower is the value the better will be performing the classifier.
    
    \item \textbf{Precision:} It is one of the most usual evaluation metrics in standard classification. It indicates the proportion of predicted positives which are truly positives, so the higher the precision the better will be performing the classifier.
    
    \item \textbf{F-measure:} Precision accounts as errors only false positives, so it is usual to include some other metric which also takes into account false negatives. F-measure is the harmonic mean of Precision and Recall (\ref{Recall}), being this last metric an indicator of the amount of false positives. Therefore, F-measure offers a more broad evaluation of the classifiers' performance than Precision or Recall in their own.
    
    \item \textbf{AUC:} Lastly, the Area Under the ROC Curve is among the most powerful metrics when it comes to assess the performance of a classifier. It evaluates the true positive rate vs the false positive rate, being a rather strict measurement of the results.
\end{itemize}

In these equations $n$ stands for the number of samples, $k$ for the number of labels, $Y_i$ is the predicted labelset, $Z_i$ the true labelset, $\Delta$ denotes the symmetric difference, and $rank(X_i, l)$ is a function that returns the confidence degree for the label $l$ in the prediction $Z_i$ provided by the classifier for the instance $X_i$. Additional details about the datasets, the MLC algorithms and these metrics, including their location into the multilabel performance metrics taxonomy, can be found in \cite{Charte:SB-MLC}.

	     \newcommand{\opA}{\mathop{\vphantom{\sum}\mathchoice
	     				{\vcenter{\hbox{\large argmax}}}
	     				{\vcenter{\hbox{\large argmax}}}{\mathrm{argmax}}{\mathrm{argmax}}}\displaylimits}

		\begin{equation}
		HammingLoss = \frac{1}{n} \frac{1}{k} \displaystyle\sum\limits_{i=1}^{n} \lvert Y_i \Delta Z_i\rvert  \label{HL}
		\end{equation}

		\begin{equation}
		Precision = \frac{1}{n} \displaystyle\sum\limits_{i=1}^{n} \frac{\lvert Y_i \cap Z_i\rvert}{\lvert Z_i\rvert} \label{Precision}
		\end{equation}

		\begin{equation}
		Recall = \frac{1}{n} \displaystyle\sum\limits_{i=1}^{n} \frac{\lvert Y_i \cap Z_i\rvert}{\lvert Y_i\rvert} \label{Recall}
		\end{equation}

		\begin{equation}
		\textit{F-Measure} = 2 * \frac{Precision * Recall}{Precision + Recall} .\label{F1}
		\end{equation}
\\

\begin{strip}
		\begin{equation}
		\textit{RankingLoss} = \frac{1}{n} \displaystyle\sum\limits_{i=1}^{n}  \frac{1}{\lvert Y_i\rvert . \lvert\overline{Y_i}\rvert} 
				\lvert{y_a, y_b : rank(x_i, y_a) > rank(x_i, y_b), (y_a, y_b) \in Y_i \times \overline{Y_i} }\rvert
		\label{RankingLoss}
		\end{equation}

        \begin{equation}
        \begin{split}
        \textit{AUC} = \frac{\lvert\{x', x'', y', y'' : rank(x', y') \ge rank(x'', y''), (x', y') \in S^+ , (x'', y'') \in S^-  \}\rvert}{\lvert S^+\rvert . \lvert S^-\rvert}, \\
         S^+ = \{ (x_i, y) | y \in Y_i\}, S^- = \{ (x_i, y) | y \notin Y_i\}
        \label{MicroAUC}
        \end{split}
        \end{equation}
\end{strip}

The results obtained from each configuration run have been collected into five tables, Table \ref{Table:HammingLoss} to Table \ref{Table:AUC}, one for each evaluation metric. The title of the table indicates the name of the metric, as well as if it is a measurement to be minimized ($\downarrow$) or maximized ($\uparrow$). 

All tables have the same structure. The three resampling algorithms appear as columns, with each one of their six configurations as subcolumns. The column dubbed as \texttt{Base} corresponds to the results produced by the resampling method in its own, while the columns entitled \texttt{H$_{nn}$} come from the hybrid version with the five thresholds previously enumerated (see diagram in Fig. \ref{Hybridization}), being $nn$ the corresponding threshold. The datasets, grouped by classifier, appear as rows. Therefore, nine subgroups can be easily identified inside each table, according to preprocessing and classification algorithms.

The configuration with the best performance for each MLD inside each subgroup has been highlighted in bold. This way it is easy to check if for a given case the hybridization has been able to improve the base resampling results or not. If there are several configurations reaching the best value, all of them are highlighted. If there is a tie for all configurations, none is emphasized. It must be noted that in many cases all hybrid configurations improve the base result, but only the best performer is pointed out.

For some configurations the MULAN framework \cite{MULAN} was not able to provide some evaluation metrics. These cases appear as dashes in the tables.

\subsection{Analysis of Results}
Here, the obtained results are going to be analyzed according to several criteria. Firstly, the focus will be in the evaluation metric, then in the resampling method, further in the classifier, and lastly in individual MLDs.

\begin{itemize}
    \item \textbf{Hamming Loss:}  Looking at Table \ref{Table:HammingLoss} it can be verified that the hybrid versions are able to improve results for the BR and LP classifiers, having between 8 and 6 best cases out of 10. There is not a clear optimum threshold, although for LP the \texttt{H1} configuration, which corresponds to the lower cut point, gathers more best values than the others. The scenario for ML-kNN is completely different, since the base resampling achieves at least 6 out of 10 best performances.
\end{itemize}

\setlength{\tabcolsep}{2pt}
\begin{landscape}
\def\arraystretch{1.1}
\centering
\begin{tabular}{lrrrrrr|rrrrrr|rrrrrr}
  \hline
Classifier/ & \multicolumn{6}{c|}{ML-ROS} & \multicolumn{6}{c|}{MLeNN} & \multicolumn{6}{c}{MLSMOTE} \\ 
Dataset & 
          \multicolumn{1}{c}{Base} & \multicolumn{1}{c}{H$_{0.25}$} & \multicolumn{1}{c}{H$_{0.37}$} & \multicolumn{1}{c}{H$_{0.50}$} & \multicolumn{1}{c}{H$_{0.62}$} & \multicolumn{1}{c|}{H$_{0.75}$} & 
          \multicolumn{1}{c}{Base} & \multicolumn{1}{c}{H$_{0.25}$} & \multicolumn{1}{c}{H$_{0.37}$} & \multicolumn{1}{c}{H$_{0.50}$} & \multicolumn{1}{c}{H$_{0.62}$} & \multicolumn{1}{c|}{H$_{0.75}$} & 
          \multicolumn{1}{c}{Base} & \multicolumn{1}{c}{H$_{0.25}$} & \multicolumn{1}{c}{H$_{0.37}$} & \multicolumn{1}{c}{H$_{0.50}$} & \multicolumn{1}{c}{H$_{0.62}$} & \multicolumn{1}{c}{H$_{0.75}$} \\

  \toprule
  \Large{BR} \\ 
  cal500 & 
  0.2053 & \textbf{0.1736} & 0.1787 & 0.1827 & 0.1873 & 0.1936 & 
  0.1660 & 0.1521 & \textbf{0.1514} & 0.1559 & 0.1573 & 0.1587 & 
  0.1611 & \textbf{0.1457} & \textbf{0.1457} & 0.1476 & 0.1510 & 0.1547 \\ 
  
  chess & 
  0.0109 & 0.0106 & 0.0106 & 0.0106 & \textbf{0.0105} & 0.0106 & 
  0.0112 & \textbf{0.0106} & 0.0107 & 0.0107 & \textbf{0.0106} & 0.0107 & 
  0.0109 & 0.0106 & 0.0106 & 0.0106 & \textbf{0.0105} & 0.0106 \\ 
  
   corel16k & 
  0.0201 & \textbf{0.0188} & \textbf{0.0188} & \textbf{0.0188} & 0.0189 & 0.0192 & 
  0.0199 & \textbf{0.0190} & 0.0191 & 0.0191 & 0.0192 & 0.0193 & 
  0.0197 & \textbf{0.0187} & \textbf{0.0187} & \textbf{0.0187} & 0.0188 & 0.0190 \\ 
  
   corel5k & 
  0.0098 & \textbf{0.0094} & \textbf{0.0094} & \textbf{0.0094} & \textbf{0.0094} & 0.0095 & 
  0.0111 & \textbf{0.0097} & \textbf{0.0097} & \textbf{0.0097} & \textbf{0.0097} & 0.0101 & 
  0.0098 & \textbf{0.0094} & \textbf{0.0094} & \textbf{0.0094} & \textbf{0.0094} & 0.0095 \\ 
  
   delicious & 
  0.0186 & 0.0190 & 0.0188 & 0.0186 & 0.0185 & \textbf{0.0184} & 
  \textbf{0.0188} & 0.0193 & 0.0193 & 0.0192 & 0.0191 & 0.0190 & 
  0.0186 & 0.0190 & 0.0188 & 0.0186 & 0.0185 & \textbf{0.0184} \\ 
  
   enron & 
  \textbf{0.0531} & 0.0635 & 0.0603 & 0.0568 & 0.0538 & 0.0537 & 
  0.0674 & 0.0720 & \textbf{0.0673} & 0.0702 & 0.0704 & 0.0695 & 
  \textbf{0.0507} & 0.0622 & 0.0580 & 0.0545 & 0.0527 & 0.0524 \\ 
  
   mediamill & 
  0.0334 & 0.0388 & 0.0355 & 0.0334 & \textbf{0.0318} & 0.0319 & 
  0.0468 & \textbf{0.0434} & \textbf{0.0434} & 0.0435 & 0.0436 & 0.0438 & 
  0.0334 & 0.0388 & 0.0355 & 0.0334 & \textbf{0.0318} & 0.0319 \\ 
  
   medical & 
  \textbf{0.0111} & 0.0142 & 0.0142 & 0.0142 & 0.0142 & 0.0142 & 
  \textbf{0.0248} & 0.0305 & 0.0305 & 0.0305 & 0.0305 & 0.0305 & 
  \textbf{0.0104} & 0.0140 & 0.0140 & 0.0140 & 0.0140 & 0.0140 \\ 
  
  tmc2007 & 
  \textbf{0.0576} & 0.0937 & 0.0845 & 0.0754 & 0.0711 & 0.0674 & 
  \textbf{0.0961} & 0.1114 & 0.1075 & 0.1089 & 0.1084 & 0.1095 & 
  \textbf{0.0550} & 0.0907 & 0.0819 & 0.0714 & 0.0676 & 0.0626 \\ 
  
  yeast & 
  0.2778 & 0.2806 & \textbf{0.2714} & 0.2737 & 0.2741 & 0.2749 & 
  \textbf{0.2995} & 0.3374 & 0.3568 & 0.3670 & 0.3494 & 0.3209 & 
  0.2467 & 0.2998 & 0.2928 & 0.2638 & 0.2464 & \textbf{0.2264} \\ 
  
  \midrule \Large{LP} \\ 
  cal500 & 
  0.2038 & \textbf{0.1751} & 0.1772 & 0.1770 & 0.1825 & 0.1893 & 
  0.2048 & \textbf{0.1687} & 0.1730 & 0.1776 & 0.1828 & 0.1914 & 
  0.2009 & \textbf{0.1725} & 0.1772 & 0.1824 & 0.1850 & 0.1908 \\ 
  
  chess & 
  0.0175 & \textbf{0.0108} & 0.0109 & 0.0116 & 0.0128 & 0.0145 & 
  0.0208 & \textbf{0.0107} & \textbf{0.0107} & \textbf{0.0107} & 0.0109 & 0.0117 & 
  0.0175 & \textbf{0.0108} & 0.0109 & 0.0116 & 0.0128 & 0.0145 \\ 
  
  corel16k & 
  0.0324 & \textbf{0.0210} & 0.0220 & 0.0234 & 0.0254 & 0.0277 & 
  0.0366 & \textbf{0.0259} & 0.0260 & 0.0272 & 0.0283 & 0.0295 & 
  0.0320 & \textbf{0.0204} & 0.0216 & 0.0232 & 0.0251 & 0.0274 \\ 
  
  corel5k & 
  0.0166 & \textbf{0.0094} & 0.0095 & 0.0102 & 0.0112 & 0.0123 & 
  0.0176 & 0.0124 & \textbf{0.0123} & 0.0125 & 0.0126 & 0.0137 & 
  0.0166 & \textbf{0.0094} & 0.0095 & 0.0102 & 0.0112 & 0.0123 \\
   
  delicious & 
  0.0299 & \textbf{0.0228} & 0.0240 & 0.0255 & 0.0267 & 0.0279 & 
  0.0307 & \textbf{0.0206} & 0.0217 & 0.0233 & 0.0251 & 0.0270 & 
  0.0299 & \textbf{0.0228} & 0.0240 & 0.0255 & 0.0267 & 0.0279 \\ 
  
  enron & 
  0.0726 & 0.0632 & 0.0634 & \textbf{0.0629} & 0.0652 & 0.0656 & 
  0.0978 & \textbf{0.0842} & 0.0894 & 0.0910 & 0.0926 & 0.0889 & 
  0.0725 & 0.0631 & 0.0634 & \textbf{0.0625} & 0.0646 & 0.0662 \\
   
  mediamill & 
  0.0423 & 0.0428 & 0.0414 & 0.0413 & \textbf{0.0402} & 0.0408 & 
  0.0562 & \textbf{0.0435} & 0.0436 & 0.0437 & 0.0439 & 0.0446 & 
  0.0423 & 0.0428 & 0.0414 & 0.0413 & \textbf{0.0402} & 0.0408 \\ 
  
  medical & 
  \textbf{0.0131} & 0.0162 & 0.0162 & 0.0162 & 0.0162 & 0.0162 &
  \textbf{0.0432} & 0.0493 & 0.0493 & 0.0493 & 0.0493 & 0.0493 & 
  \textbf{0.0130} & 0.0161 & 0.0161 & 0.0161 & 0.0161 & 0.0161 \\ 
  
  tmc2007 & 
  \textbf{0.0697} & 0.0944 & 0.0906 & 0.0869 & 0.0827 & 0.0788 & 
  0.1382 & \textbf{0.1353} & 0.1370 & 0.1372 & 0.1373 & 0.1385 & 
  \textbf{0.0708} & 0.0927 & 0.0888 & 0.0849 & 0.0809 & 0.0766 \\ 
  
  yeast & 
  0.2902 & 0.3161 & 0.3076 & 0.3078 & 0.2914 & \textbf{0.2897} & 
  \textbf{0.3254} & 0.3495 & 0.3699 & 0.3765 & 0.3693 & 0.3574 & 
  0.2815 & 0.3079 & 0.2987 & 0.2944 & 0.2783 & \textbf{0.2713} \\
   
  \midrule \Large{ML-kNN} \\ cal500 & 
  0.1877 & \textbf{0.1598} & 0.1615 & 0.1639 & 0.1672 & 0.1763 & 
  \textbf{0.1397} & 0.1497 & 0.1490 & 0.1494 & 0.1478 & 0.1439 & 
  \textbf{0.1387} & 0.1469 & 0.1466 & 0.1452 & 0.1432 & 0.1415 \\ 
  
  chess & 
  0.0106 & 0.0106 & 0.0106 & 0.0106 & 0.0106 & 0.0106 & 
  0.0106 & 0.0106 & 0.0106 & 0.0106 & 0.0106 & 0.0106 & 
  0.0106 & 0.0106 & 0.0106 & 0.0106 & 0.0106 & 0.0106 \\
   
  corel16k & 
  0.0189 & \textbf{0.0187} & \textbf{0.0187} & \textbf{0.0187} & 0.0188 & 0.0188 & 
  0.0187 & 0.0187 & 0.0187 & 0.0187 & 0.0187 & 0.0187 & 
  \textbf{0.0186} & 0.0187 & 0.0187 & 0.0186 & 0.0187 & \textbf{0.0186} \\ 
  
  corel5k & 
  0.0094 & 0.0094 & 0.0094 & 0.0094 & 0.0094 & 0.0094 & 
  0.0096 & \textbf{0.0094} & \textbf{0.0094} & \textbf{0.0094} & \textbf{0.0094} & 0.0095 & 
  0.0094 & 0.0094 & 0.0094 & 0.0094 & 0.0094 & 0.0094 \\ 
  
  delicious & 
  \textbf{0.0184} & 0.0191 & 0.0190 & 0.0189 & 0.0188 & 0.0186 & 
  \textbf{0.0185} & 0.0192 & 0.0192 & 0.0192 & 0.0191 & 0.0190 & 
  \textbf{0.0184} & 0.0191 & 0.0190 & 0.0189 & 0.0188 & 0.0186 \\ 
  
  enron & 
  \textbf{0.0528} & 0.0635 & 0.0627 & 0.0614 & 0.0590 & 0.0563 & 
  \textbf{0.0611} & 0.0644 & 0.0647 & 0.0645 & 0.0633 & 0.0635 & 
  \textbf{0.0518} & 0.0633 & 0.0625 & 0.0613 & 0.0587 & 0.0560 \\ 
  
  mediamill & 
  \textbf{0.0282} & 0.0381 & 0.0370 & 0.0346 & 0.0309 & 0.0295 & 
  \textbf{0.0372} & 0.0433 & 0.0433 & 0.0433 & 0.0433 & 0.0434 & 
  \textbf{0.0282} & 0.0381 & 0.0370 & 0.0346 & 0.0309 & 0.0295 \\ 
  
  medical &
  \textbf{0.0174} & 0.0182 & 0.0182 & 0.0182 & 0.0182 & 0.0182 & 
  0.0308 & \textbf{0.0281} & \textbf{0.0281} & \textbf{0.0281} & \textbf{0.0281} & \textbf{0.0281} & 
  \textbf{0.0157} & 0.0173 & 0.0173 & 0.0173 & 0.0173 & 0.0173 \\ 
  
  tmc2007 & 
  \textbf{0.0680} & 0.0960 & 0.0920 & 0.0856 & 0.0805 & 0.0726 & 
  \textbf{0.1018} & 0.1141 & 0.1095 & 0.1120 & 0.1101 & 0.1110 & 
  \textbf{0.0653} & 0.0975 & 0.0922 & 0.0862 & 0.0827 & 0.0693 \\ 
  
  yeast & 
  \textbf{0.2438} & 0.2812 & 0.2702 & 0.2654 & 0.2480 & 0.2477 & 
  \textbf{0.2322} & 0.2849 & 0.3142 & 0.3179 & 0.3104 & 0.2760 & 
  \textbf{0.1923} & 0.2752 & 0.2787 & 0.2707 & 0.2270 & 0.1995 \\ 
   \bottomrule
\end{tabular}
\captionof{table}{Hamming Loss ($\downarrow$)}
\label{Table:HammingLoss}
\end{landscape}

\begin{landscape}
\centering
\def\arraystretch{1.1}
\begin{tabular}{lrrrrrr|rrrrrr|rrrrrr}
  \hline
Classifier/ & \multicolumn{6}{c|}{ML-ROS} & \multicolumn{6}{c|}{MLeNN} & \multicolumn{6}{c}{MLSMOTE} \\ 
Dataset & 
          \multicolumn{1}{c}{Base} & \multicolumn{1}{c}{H$_{0.25}$} & \multicolumn{1}{c}{H$_{0.37}$} & \multicolumn{1}{c}{H$_{0.50}$} & \multicolumn{1}{c}{H$_{0.62}$} & \multicolumn{1}{c|}{H$_{0.75}$} & 
          \multicolumn{1}{c}{Base} & \multicolumn{1}{c}{H$_{0.25}$} & \multicolumn{1}{c}{H$_{0.37}$} & \multicolumn{1}{c}{H$_{0.50}$} & \multicolumn{1}{c}{H$_{0.62}$} & \multicolumn{1}{c|}{H$_{0.75}$} & 
          \multicolumn{1}{c}{Base} & \multicolumn{1}{c}{H$_{0.25}$} & \multicolumn{1}{c}{H$_{0.37}$} & \multicolumn{1}{c}{H$_{0.50}$} & \multicolumn{1}{c}{H$_{0.62}$} & \multicolumn{1}{c}{H$_{0.75}$} \\

  \toprule
  \Large{BR} \\ 
  cal500 & 
  0.3827 & \textbf{0.2912} & 0.3015 & 0.3094 & 0.3166 & 0.3190 & 
  0.3276 & 0.2496 & \textbf{0.2120} & 0.2495 & 0.2618 & 0.2850 & 
  0.2968 & \textbf{0.1816} & 0.1831 & 0.1993 & 0.2176 & 0.2408 \\ 
  
  chess & 
  0.1933 & 0.1710 & 0.1715 & \textbf{0.1709} & 0.1751 & 0.1753 & 
  \textbf{0.2666} & 0.6236 & 0.6189 & 0.6336 & 0.6375 & 0.5857 & 
  0.1933 & 0.1710 & 0.1715 & \textbf{0.1709} & 0.1751 & 0.1753 \\ 
  
  corel16k & 
  0.1861 & 0.1883 & 0.1870 & 0.1845 & \textbf{0.1836} & 0.1839 & 
  \textbf{0.3059} & 0.6499 & 0.6919 & 0.7072 & 0.6815 & 0.6068 & 
  0.1891 & 0.1889 & 0.1876 & 0.1850 & \textbf{0.1848} & 0.1857 \\ 
  
  corel5k & 
  0.1457 & 0.1421 & 0.1415 & \textbf{0.1407} & 0.1413 & 0.1418 & 
  \textbf{0.3105} & 0.6848 & 0.7070 & 0.7048 & 0.6994 & 0.6483 & 
  0.1457 & 0.1421 & 0.1415 & \textbf{0.1407} & 0.1413 & 0.1418 \\ 
  
  delicious & 
  0.1786 & \textbf{0.1628} & 0.1631 & 0.1629 & 0.1635 & 0.1663 & 
  \textbf{0.1824} & 0.3900 & 0.2490 & 0.2021 & 0.1849 & 0.1833 & 
  0.1786 & \textbf{0.1628} & 0.1631 & 0.1629 & 0.1635 & 0.1663 \\ 
  
  enron & 
  0.1682 & 0.1278 & \textbf{0.1273} & 0.1345 & 0.1424 & 0.1522 & 
  \textbf{0.2316} & 0.3411 & 0.3765 & 0.3561 & 0.3239 & 0.3859 & 
  0.1701 & 0.1281 & \textbf{0.1256} & 0.1361 & 0.1448 & 0.1535 \\ 
  
  mediamill & 
  0.1731 & \textbf{0.0648} & 0.0672 & 0.0717 & 0.0807 & 0.1029 & 
  \textbf{0.2760} & 0.6359 & 0.6381 & 0.6423 & 0.6400 & 0.6370 & 
  0.1731 & \textbf{0.0648} & 0.0672 & 0.0717 & 0.0807 & 0.1029 \\ 
  
  medical & 
  \textbf{0.0611} & 0.0641 & 0.0641 & 0.0641 & 0.0641 & 0.0641 & 
  \textbf{0.3212} & 0.6505 & 0.6505 & 0.6505 & 0.6505 & 0.6505 & 
  \textbf{0.0663} & 0.0678 & 0.0678 & 0.0678 & 0.0678 & 0.0678 \\ 
  
  tmc2007 & 
  0.1025 & 0.1123 & 0.1098 & 0.0894 & \textbf{0.0849} & 0.0867 & 
  \textbf{0.1938} & 0.7044 & 0.3012 & 0.3367 & 0.3611 & 0.3570 & 
  \textbf{0.1098} & 0.1399 & 0.1385 & 0.1292 & 0.1268 & 0.1150 \\ 
  
  yeast & 
  0.3269 & \textbf{0.2717} & 0.2863 & 0.2922 & 0.3014 & 0.2989 & 
  \textbf{0.3467} & 0.4260 & 0.4773 & 0.4999 & 0.4889 & 0.3639 & 
  0.3131 & \textbf{0.2056} & 0.2159 & 0.2277 & 0.2451 & 0.2555 \\ 
  
  \midrule\Large{LP} \\
  cal500 & 
  \textbf{0.4907} & 0.5753 & 0.5694 & 0.5600 & 0.5589 & 0.5270 & 
  0.6551 & 0.6368 & 0.6397 & \textbf{0.6356} & 0.6442 & 0.6535 & 
  0.6539 & \textbf{0.6413} & 0.6440 & 0.6441 & 0.6516 & 0.6457 \\ 
  
  chess & 
  0.4736 & 0.4530 & 0.4524 & 0.4502 & \textbf{0.4472} & 0.4583 & 
  0.4729 & 0.4535 & 0.4534 & \textbf{0.4527} & 0.4531 & 0.4537 & 
  0.4736 & 0.4530 & 0.4524 & 0.4502 & \textbf{0.4472} & 0.4583 \\ 
  
  corel16k & 
  0.4583 & \textbf{0.4377} & 0.4390 & 0.4405 & 0.4425 & 0.4472 & 
  0.4626 & 0.4356 & 0.4354 & 0.4350 & 0.4353 & \textbf{0.4344} & 
  0.4657 & \textbf{0.4378} & 0.4393 & 0.4424 & 0.4425 & 0.4462 \\ 
  
  corel5k & 
  0.7474 & 0.7530 & 0.7523 & 0.7508 & 0.7465 & \textbf{0.7448} & 
  \textbf{0.7452} & 0.7521 & 0.7522 & 0.7522 & 0.7521 & 0.7497 & 
  0.7474 & 0.7530 & 0.7523 & 0.7508 & 0.7465 & \textbf{0.7448} \\ 
  
  delicious & 
  0.5476 & \textbf{0.4884} & 0.4964 & 0.5064 & 0.5149 & 0.5231 & 
  0.5423 & \textbf{0.4716} & 0.4761 & 0.4821 & 0.4893 & 0.5035 & 
  0.5476 & \textbf{0.4884} & 0.4964 & 0.5064 & 0.5149 & 0.5231 \\ 
  
  enron & 
  0.5406 & 0.5632 & 0.5606 & 0.5503 & 0.5403 & \textbf{0.5346} & 
  \textbf{0.6079} & 0.6179 & 0.6259 & 0.6376 & 0.6279 & 0.6222 & 
  0.5378 & 0.5670 & 0.5614 & 0.5574 & 0.5488 & \textbf{0.5358} \\ 
  
  mediamill & 
  \textbf{0.3375} & 0.4020 & 0.3721 & 0.3644 & 0.3489 & 0.3427 & 
  0.4188 & 0.4182 & 0.4182 & 0.4182 & \textbf{0.4180} & 0.4181 & 
  \textbf{0.3375} & 0.4020 & 0.3721 & 0.3644 & 0.3489 & 0.3427 \\ 
  
  medical & 
  \textbf{0.1347} & 0.2130 & 0.2130 & 0.2130 & 0.2130 & 0.2130 & 
  \textbf{0.5021} & 0.5493 & 0.5493 & 0.5493 & 0.5493 & 0.5493 & 
  \textbf{0.1365} & 0.2196 & 0.2196 & 0.2196 & 0.2196 & 0.2196 \\ 
  
  chess & 
  0.4736 & 0.4530 & 0.4524 & 0.4502 & \textbf{0.4472} & 0.4583 & 
  0.4729 & 0.4535 & 0.4534 & \textbf{0.4527} & 0.4531 & 0.4537 & 
  0.4736 & 0.4530 & 0.4524 & 0.4502 & \textbf{0.4472} & 0.4583 \\ 
  
  tmc2007 & 
  \textbf{0.2960} & 0.5574 & 0.5301 & 0.4906 & 0.4434 & 0.3980 & 
  \textbf{0.4128} & 0.6063 & 0.5490 & 0.5935 & 0.5946 & 0.5994 & 
  \textbf{0.3231} & 0.5718 & 0.5423 & 0.5022 & 0.4560 & 0.4095 \\ 
  
  yeast & 
  \textbf{0.3843} & 0.4746 & 0.4545 & 0.4493 & 0.4222 & 0.4065 & 
  \textbf{0.4373} & 0.4918 & 0.5182 & 0.5280 & 0.5363 & 0.4747 & 
  \textbf{0.4000} & 0.5054 & 0.4772 & 0.4571 & 0.4320 & 0.4167 \\ 
  
  \midrule\Large{ML-kNN} \\
  cal500 & 
  0.2191 & \textbf{0.2190} & 0.2197 & 0.2202 & 0.2205 & 0.2201 & 
  \textbf{0.1890} & 0.2539 & 0.2083 & 0.2154 & 0.1957 & 0.1929 & 
  \textbf{0.1831} & 0.1857 & 0.1849 & 0.1848 & 0.1841 & 0.1834 \\ 
  
  chess & 
  \textbf{0.1543} & 0.1574 & 0.1568 & 0.1562 & 0.1556 & 0.1555 & 
  \textbf{0.2509} & 0.6262 & 0.6231 & 0.6400 & 0.6383 & 0.5876 & 
  \textbf{0.1543} & 0.1574 & 0.1568 & 0.1562 & 0.1556 & 0.1555 \\ 
  
  corel16k & 
  \textbf{0.1730} & 0.1760 & 0.1755 & 0.1748 & 0.1743 & 0.1739 & 
  \textbf{0.2971} & 0.6483 & 0.6904 & 0.7023 & 0.6732 & 0.5986 & 
  \textbf{0.1718} & 0.1751 & 0.1745 & 0.1741 & 0.1732 & 0.1729 \\ 
  
  corel5k & 
  0.1337 & 0.1340 & 0.1336 & 0.1337 & \textbf{0.1334} & 0.1336 & 
  \textbf{0.2963} & 0.6808 & 0.7044 & 0.7022 & 0.6958 & 0.6379 & 
  0.1337 & 0.1340 & 0.1336 & 0.1337 & \textbf{0.1334} & 0.1336 \\ 
  
  delicious & 
  \textbf{0.1261} & 0.1297 & 0.1291 & 0.1285 & 0.1279 & 0.1271 & 
  \textbf{0.1285} & 0.3597 & 0.2202 & 0.1692 & 0.1479 & 0.1365 & 
  \textbf{0.1261} & 0.1297 & 0.1291 & 0.1285 & 0.1279 & 0.1271 \\ 
  
  enron & 
  \textbf{0.0929} & 0.0944 & 0.0941 & 0.0934 & 0.0932 & 0.0937 & 
  \textbf{0.1727} & 0.3180 & 0.3084 & 0.2869 & 0.2544 & 0.3180 & 
  \textbf{0.0921} & 0.0938 & 0.0936 & 0.0930 & 0.0927 & 0.0931 \\ 
  
  mediamill & 
  \textbf{0.0372} & 0.0383 & 0.0378 & 0.0376 & 0.0373 & 0.0373 & 
  \textbf{0.1085} & 0.6407 & 0.6422 & 0.6485 & 0.6484 & 0.6493 & 
  \textbf{0.0372} & 0.0383 & 0.0378 & 0.0376 & 0.0373 & 0.0373 \\ 
  
  medical & 
  \textbf{0.0445} & 0.0469 & 0.0469 & 0.0469 & 0.0469 & 0.0469 & 
  \textbf{0.3179} & 0.7105 & 0.7105 & 0.7105 & 0.7105 & 0.7105 & 
  \textbf{0.0404} & 0.0452 & 0.0452 & 0.0452 & 0.0452 & 0.0452 \\ 
  
  tmc2007 & 
  \textbf{0.0614} & 0.0652 & 0.0648 & 0.0638 & 0.0642 & 0.0639 & 
  \textbf{0.1510} & 0.7074 & 0.3214 & 0.4262 & 0.4345 & 0.4068 & 
  \textbf{0.0586} & 0.0637 & 0.0635 & 0.0616 & 0.0620 & 0.0611 \\ 
  
  yeast & 
  \textbf{0.1990} & 0.2020 & 0.2021 & 0.2030 & 0.2018 & 0.2029 & 
  \textbf{0.2042} & 0.3884 & 0.4955 & 0.5441 & 0.5448 & 0.2919 & 
  \textbf{0.1650} & 0.1815 & 0.1766 & 0.1756 & 0.1693 & 0.1668 \\ 
   \bottomrule
\end{tabular}
\captionof{table}{Ranking Loss ($\downarrow$)}
\label{Table:RankingLoss}
\end{landscape}

\begin{landscape}
\def\arraystretch{1.1}
\centering
\begin{tabular}{lrrrrrr|rrrrrr|rrrrrr}
  \hline
Classifier/ & \multicolumn{6}{c|}{ML-ROS} & \multicolumn{6}{c|}{MLeNN} & \multicolumn{6}{c}{MLSMOTE} \\ 
Dataset & 
          \multicolumn{1}{c}{Base} & \multicolumn{1}{c}{H$_{0.25}$} & \multicolumn{1}{c}{H$_{0.37}$} & \multicolumn{1}{c}{H$_{0.50}$} & \multicolumn{1}{c}{H$_{0.62}$} & \multicolumn{1}{c|}{H$_{0.75}$} & 
          \multicolumn{1}{c}{Base} & \multicolumn{1}{c}{H$_{0.25}$} & \multicolumn{1}{c}{H$_{0.37}$} & \multicolumn{1}{c}{H$_{0.50}$} & \multicolumn{1}{c}{H$_{0.62}$} & \multicolumn{1}{c|}{H$_{0.75}$} & 
          \multicolumn{1}{c}{Base} & \multicolumn{1}{c}{H$_{0.25}$} & \multicolumn{1}{c}{H$_{0.37}$} & \multicolumn{1}{c}{H$_{0.50}$} & \multicolumn{1}{c}{H$_{0.62}$} & \multicolumn{1}{c}{H$_{0.75}$} \\
  \toprule
  \Large{BR} \\ 
  cal500 & 
  0.3404 & \textbf{0.3444} & 0.3289 & 0.3299 & 0.3409 & 0.3357 & 
  \textbf{0.4290} & 0.2294 & 0.2983 & 0.2738 & 0.3640 & 0.4219 &
  0.4458 & \textbf{0.8455} & 0.7563 & 0.5952 & 0.5140 & 0.4719 \\ 
  
  chess & 
  0.5154 & 0.7066 & 0.6163 & 0.4816 & 0.5354 & 0.5422 & 
  0.4856 & - & - & - & - & - & 
  0.5154 & 0.7066 & 0.6163 & 0.4816 & 0.5354 & 0.5422 \\ 
  
  corel16k & 
  0.3289 & 0.3391 & 0.3549 & \textbf{0.3977} & 0.3704 & 0.3595 & 
  \textbf{0.2687} & 0.0723 & 0.0711 & 0.0951 & 0.1056 & 0.1165 & 
  0.3578 & \textbf{0.4866} & 0.4577 & 0.4787 & 0.4306 & 0.4000 \\ 
  
  corel5k & 
  0.3687 & \textbf{0.6013} & 0.4951 & 0.5025 & 0.4673 & 0.4270 & 
  \textbf{0.1562} & 0.0137 & 0.0140 & 0.0140 & 0.0140 & 0.0364 &
  0.3687 & \textbf{0.6013} & 0.4951 & 0.5025 & 0.4673 & 0.4270 \\ 
  
  delicious & 
  0.5416 & \textbf{0.7699} & 0.7357 & 0.7051 & 0.6611 & 0.6128 & 
  0.5356 & 0.4904 & 0.5026 & 0.5434 & 0.5676 & \textbf{0.5852} & 
  0.5416 & \textbf{0.7699} & 0.7357 & 0.7051 & 0.6611 & 0.6128 \\ 
  
  enron & 
  0.6418 & 0.5496 & 0.6656 & 0.6708 & \textbf{0.6760} & 0.6582 & 
  \textbf{0.4585} & 0.2846 & 0.4170 & 0.3973 & 0.3900 & 0.3429 & 
  0.6565 & 0.6666 & \textbf{0.7477} & 0.7031 & 0.6989 & 0.6611 \\ 
  
  mediamill & 
  0.6517 & \textbf{0.9061} & 0.8480 & 0.8060 & 0.7670 & 0.7179 & 
  \textbf{0.4605} & 0.3667 & 0.2986 & 0.2628 & 0.2120 & 0.1470 & 
  0.6517 & \textbf{0.9061} & 0.8480 & 0.8060 & 0.7670 & 0.7179 \\ 
  
  medical & 
  0.8593 & \textbf{0.8698} & \textbf{0.8698} & \textbf{0.8698} & \textbf{0.8698} & \textbf{0.8698} & 
  \textbf{0.6414} & 0.1126 & 0.1126 & 0.1126 & 0.1126 & 0.1126 & 
  0.8700 & \textbf{0.8714} & \textbf{0.8714} & \textbf{0.8714} & \textbf{0.8714} & \textbf{0.8714} \\ 
  
  tmc2007 & 
  0.7533 & 0.7546 & 0.8012 & \textbf{0.8053} & 0.7925 & 0.7727 & 
  \textbf{0.5597} & 0.2872 & 0.4209 & 0.3403 & 0.3520 & 0.3403 & 
  0.7768 & \textbf{0.9130} & 0.8761 & 0.8457 & 0.8229 & 0.7971 \\ 
  
  yeast & 
  0.5536 & 0.5628 & \textbf{0.5781} & 0.5725 & 0.5604 & 0.5579 & 
  \textbf{0.5193} & 0.3857 & 0.3179 & 0.2706 & 0.2704 & 0.4218 & 
  0.6098 & 0.6075 & 0.6075 & \textbf{0.6690} & 0.6518 & 0.6649 \\ 
  
  \midrule\Large{LP} \\
  cal500 & 
  \textbf{0.3261} & 0.1557 & 0.1586 & 0.1771 & 0.1875 & 0.2372 & 
  \textbf{0.3227} & 0.1220 & 0.1452 & 0.1483 & 0.1737 & 0.2399 & 
  \textbf{0.3304} & 0.1839 & 0.2132 & 0.2354 & 0.2625 & 0.3024 \\ 
  
  chess & 
  0.1676 & 0.1042 & \textbf{0.2108} & 0.1760 & 0.1919 & 0.1959 & 
  \textbf{0.0693} & 0.0268 & 0.0268 & 0.0268 & 0.0268 & 0.0268 & 
  0.1676 & 0.1042 & \textbf{0.2108} & 0.1760 & 0.1919 & 0.1959 \\ 
  
  corel16k & 
  0.1268 & 0.1077 & 0.1132 & 0.1143 & 0.1206 & \textbf{0.1276} & 
  \textbf{0.0713} & 0.0189 & 0.0180 & 0.0203 & 0.0222 & 0.0247 & 
  \textbf{0.1338} & 0.0994 & 0.1177 & 0.1204 & 0.1242 & 0.1285 \\ 
  
  corel5k & 
  0.1141 & 0.1471 & 0.1471 & \textbf{0.1756} & 0.1471 & 0.1386 & 
  \textbf{0.0577} & 0.0047 & 0.0047 & 0.0034 & 0.0049 & 0.0098 & 
  0.1141 & 0.1386 & 0.1386 & \textbf{0.1756} & 0.1471 & 0.1386 \\ 
  
  delicious & 
  \textbf{0.2306} & 0.2190 & 0.2026 & 0.1965 & 0.1977 & 0.2043 & 
  \textbf{0.2103} & 0.1206 & 0.1143 & 0.1191 & 0.1209 & 0.1419 & 
  \textbf{0.2306} & 0.2190 & 0.2026 & 0.1965 & 0.1977 & 0.2043 \\ 
  
  enron & 
  0.4602 & \textbf{0.5914} & 0.5443 & 0.5500 & 0.4987 & 0.4887 & 
  \textbf{0.2974} & 0.1585 & 0.1756 & 0.1639 & 0.1754 & 0.1507 & 
  0.4584 & \textbf{0.6199} & 0.5793 & 0.5603 & 0.5052 & 0.4864 \\ 
  
  mediamill & 
  0.5466 & 0.6144 & \textbf{0.6216} & 0.5896 & 0.5876 & 0.5655 & 
  \textbf{0.3385} & 0.2468 & 0.2246 & 0.1585 & 0.1386 & 0.0835 & 
  0.5466 & 0.6144 & 0.6216 & 0.5896 & 0.5876 & 0.5655 \\ 
  
  medical & 
  0.7902 & \textbf{0.7943} & \textbf{0.7943} & \textbf{0.7943} & \textbf{0.7943} & \textbf{0.7943} & 
  \textbf{0.2038} & 0.0127 & 0.0127 & 0.0127 & 0.0127 & 0.0127 & 
  0.7893 & \textbf{0.7938} & \textbf{0.7938} & \textbf{0.7938} & \textbf{0.7938} & \textbf{0.7938} \\ 
  
  tmc2007 & 
  0.6816 & 0.6716 & \textbf{0.6817} & 0.6739 & 0.6691 & 0.6684 & 
  \textbf{0.3744} & 0.1308 & 0.2107 & 0.1562 & 0.1626 & 0.1505 & 
  0.6851 & \textbf{0.7366} & 0.7308 & 0.7110 & 0.6945 & 0.6885 \\ 
  
  yeast & 
  \textbf{0.5250} & 0.4018 & 0.4312 & 0.4374 & 0.4740 & 0.4841 & 
  \textbf{0.4702} & 0.3097 & 0.2110 & 0.1843 & 0.1812 & 0.3094 & 
  \textbf{0.5361} & 0.3769 & 0.4594 & 0.4747 & 0.5118 & 0.5245 \\ 
  
  \midrule\Large{ML-kNN} \\
  cal500 & 
  \textbf{0.3669} & 0.2392 & 0.2346 & 0.2565 & 0.2594 & 0.2943 & 
  0.5951 & 0.6906 & 0.6906 & 0.5812 & \textbf{0.7566} & 0.7420 & 
  0.6038 & \textbf{0.8380} & 0.8361 & 0.8063 & 0.7787 & 0.6906 \\ 
  
  corel16k & 
  \textbf{0.3865} & 0.1684 & 0.3137 & 0.3533 & 0.3322 & 0.3714 & 
  \textbf{0.2539} & 0.0999 & 0.1493 & 0.1852 & 0.2398 & 0.2425 & 
  0.5403 & 0.5083 & 0.5580 & \textbf{0.6162} & 0.5861 & 0.5377 \\ 
  
  corel5k & 
  0.5839 & 0.7068 & \textbf{0.7115} & 0.6765 & 0.6367 & 0.6233 &
  0.2790 & - & - & - & - & - & 
  0.5839 & 0.7068 & \textbf{0.7115} & 0.6765 & 0.6367 & 0.6233 \\ 
  
  delicious & 
  0.6113 & \textbf{0.7810} & 0.7750 & 0.7712 & 0.7635 & 0.7241 & 
  0.6171 & 0.6280 & 0.6352 & 0.6626 & 0.6841 & \textbf{0.7015} & 
  0.6113 & \textbf{0.7810} & 0.7750 & 0.7712 & 0.7635 & 0.7241 \\ 
  
  enron & 
  0.6640 & 0.5623 & 0.6310 & 0.7066 & \textbf{0.7457} & 0.7202 & 
  0.5077 & 0.4307 & 0.3929 & 0.4698 & 0.4994 & \textbf{0.5377} & 
  0.6744 & 0.6833 & 0.7042 & \textbf{0.7606} & 0.7703 & 0.7452 \\ 
  
  mediamill & 
  0.7756 & \textbf{0.8203} & 0.7826 & 0.8196 & 0.8175 & 0.8137 & 
  \textbf{0.7353} & 0.6045 & 0.5635 & 0.5016 & 0.4677 & 0.3644 & 
  0.7756 & \textbf{0.8203} & 0.7826 & 0.8196 & 0.8175 & 0.8137 \\ 
  
  medical & 
  0.7799 & \textbf{0.8482} & \textbf{0.8482} & \textbf{0.8482} & \textbf{0.8482} & \textbf{0.8482} & 
  0.3448 & - & - & - & - & - & 
  0.8276 & \textbf{0.8654} & \textbf{0.8654} & \textbf{0.8654} & \textbf{0.8654} & \textbf{0.8654} \\ 
  
  tmc2007 & 
  0.7058 & 0.7791 & 0.7996 & \textbf{0.8100} & 0.7756 & 0.7378 & 
  \textbf{0.5058} & 0.2237 & 0.3329 & 0.2562 & 0.2677 & 0.2543 & 
  0.7261 & \textbf{0.8939} & 0.8625 & 0.8483 & 0.8224 & 0.7504 \\ 
  
  yeast & 
  0.6129 & 0.5759 & 0.6106 & 0.6198 & \textbf{0.6290} & 0.6110 & 
  \textbf{0.6327} & 0.5008 & 0.3448 & 0.2921 & 0.2716 & 0.5666 & 
  0.7256 & 0.6873 & 0.6776 & 0.6957 & 0.7391 & \textbf{0.7469} \\ 
   \bottomrule
\end{tabular}
\captionof{table}{Precision ($\uparrow$)}
\label{Table:Precision}
\end{landscape}

\begin{landscape}
\def\arraystretch{1.1}
\centering
\begin{tabular}{lrrrrrr|rrrrrr|rrrrrr}
  \hline
Classifier/ & \multicolumn{6}{c|}{ML-ROS} & \multicolumn{6}{c|}{MLeNN} & \multicolumn{6}{c}{MLSMOTE} \\ 
Dataset & 
          \multicolumn{1}{c}{Base} & \multicolumn{1}{c}{H$_{0.25}$} & \multicolumn{1}{c}{H$_{0.37}$} & \multicolumn{1}{c}{H$_{0.50}$} & \multicolumn{1}{c}{H$_{0.62}$} & \multicolumn{1}{c|}{H$_{0.75}$} & 
          \multicolumn{1}{c}{Base} & \multicolumn{1}{c}{H$_{0.25}$} & \multicolumn{1}{c}{H$_{0.37}$} & \multicolumn{1}{c}{H$_{0.50}$} & \multicolumn{1}{c}{H$_{0.62}$} & \multicolumn{1}{c|}{H$_{0.75}$} & 
          \multicolumn{1}{c}{Base} & \multicolumn{1}{c}{H$_{0.25}$} & \multicolumn{1}{c}{H$_{0.37}$} & \multicolumn{1}{c}{H$_{0.50}$} & \multicolumn{1}{c}{H$_{0.62}$} & \multicolumn{1}{c}{H$_{0.75}$} \\
  \toprule
  \Large{BR} \\ 
  cal500 & 
  \textbf{0.3317} & 0.1973 & 0.2034 & 0.2233 & 0.2528 & 0.2777 & 
  \textbf{0.3514} & 0.0876 & 0.0865 & 0.1153 & 0.1430 & 0.2162 & 
  \textbf{0.3410} & 0.0778 & 0.1006 & 0.1539 & 0.1872 & 0.2363 \\ 
  
  chess & 
  0.5805 & 0.\textbf{6092} & \textbf{0.6092} & 0.5530 & 0.5946 & 0.5839 & 
  0.5584 & - & - & - & - & - & 
  0.5805 & \textbf{0.6092} & \textbf{0.6092} & 0.5530 & 0.5946 & 0.5839 \\ 
  
  corel16k & 
  0.5148 & 0.6449 & \textbf{0.6681} & 0.6179 & 0.5948 & 0.5560 & 
  0.5267 & 0.4919 & 0.5040 & \textbf{0.5996} & 0.5912 & 0.6127 & 
  0.5229 & 0.6495 & \textbf{0.6828} & 0.6322 & 0.5888 & 0.5530 \\ 
  
  corel5k & 
  0.4728 & \textbf{0.5610} & 0.5380 & 0.5105 & 0.5043 & 0.4950 & 
  0.4433 & - & - & - & - & - & 
  0.4728 & \textbf{0.5610} & 0.5380 & 0.5105 & 0.5043 & 0.4950 \\ 
  
  delicious & 
  0.3107 & \textbf{0.3438} & 0.2672 & 0.2598 & 0.2729 & 0.2881 & 
  0.2969 & \textbf{0.4093} & 0.3713 & 0.3257 & 0.2758 & 0.2615 & 
  0.3107 & \textbf{0.3438} & 0.2672 & 0.2598 & 0.2729 & 0.2881 \\ 
  
  enron & 
  0.6033 & \textbf{0.6745} & 0.5703 & 0.5697 & 0.5806 & 0.5837 & 
  \textbf{0.5270} & 0.4517 & 0.4842 & 0.4649 & 0.4690 & 0.4704 & 
  0.6103 & \textbf{0.6584} & 0.5500 & 0.5833 & 0.5817 & 0.5908 \\ 
  
  mediamill & 
  \textbf{0.6085} & 0.4424 & 0.5337 & 0.5582 & 0.5927 & 0.6010 & 
  \textbf{0.4477} & 0.3675 & 0.3678 & 0.3688 & 0.3696 & 0.3715 & 
  \textbf{0.6085} & 0.4424 & 0.5337 & 0.5582 & 0.5927 & 0.6010 \\ 
  
  medical & 
  \textbf{0.9200} & 0.9171 & 0.9171 & 0.9171 & 0.9171 & 0.9171 & 
  0.8211 & - & - & - & - & - & 
  \textbf{0.9259} & 0.9154 & 0.9154 & 0.9154 & 0.9154 & 0.9154 \\ 
  
  tmc2007 & 
  \textbf{0.7770} & 0.6928 & 0.7087 & 0.7057 & 0.7173 & 0.7276 & 
  \textbf{0.6389} & 0.6142 & 0.5955 & 0.6110 & 0.6104 & 0.6112 & 
  \textbf{0.7808} & 0.7382 & 0.7392 & 0.7255 & 0.7329 & 0.7485 \\ 
  
  yeast & 
  \textbf{0.5776} & 0.5183 & 0.5465 & 0.5546 & 0.5686 & 0.5688 & 
  \textbf{0.5712} & 0.4642 & 0.4102 & 0.3921 & 0.4184 & 0.4811 & 
  0.6229 & 0.4412 & 0.4412 & 0.4687 & 0.5828 & \textbf{0.6326} \\ 
  
  \midrule\Large{LP} \\
  cal500 & 
  0.3217 & 0.3128 & 0.3189 & 0.3214 & 0.3089 & \textbf{0.3259} & 
  0.3213 & 0.3217 & 0.3170 & 0.3250 & 0.3275 & \textbf{0.3276} & 
  0.3230 & 0.3136 & 0.3162 & 0.3275 & \textbf{0.3394} & 0.3278 \\ 
  
  chess & 
  0.4845 & 0.5481 & 0.5481 & \textbf{0.5501} & 0.5481 & 0.5123 & 
  0.3935 & - & - & - & - & - & 
  0.4845 & 0.5481 & 0.5481 & \textbf{0.5501} & 0.5481 & 0.5123 \\ 
  
  corel16k & 
  0.4613 & \textbf{0.5423} & 0.5129 & 0.5014 & 0.4843 & 0.4863 & 
  0.3923 & 0.5083 & 0.5007 & 0.5222 & \textbf{0.5300} & 0.5244 & 
  0.4614 & \textbf{0.5262} & 0.5173 & 0.5063 & 0.4880 & 0.4821 \\ 
  
  corel5k & 
  0.4014 & 0.4678 & 0.4678 & \textbf{0.4811} & 0.4678 & 0.4544 & 
  0.3458 & 0.4022 & 0.3559 & 0.3559 & 0.3559 & \textbf{0.4479} & 
  0.4014 & 0.4678 & 0.4678 & \textbf{0.4811} & 0.4678 & 0.4544 \\ 
  
  delicious & 
  0.2609 & \textbf{0.2879} & 0.2734 & 0.2638 & 0.2619 & 0.2569 & 
  0.2444 & \textbf{0.2833} & 0.2548 & 0.2435 & 0.2327 & 0.2360 & 
  0.2609 & \textbf{0.2879} & 0.2734 & 0.2638 & 0.2619 & 0.2569 \\ 
  
  enron & 
  0.5579 & \textbf{0.8135} & 0.7468 & 0.7131 & 0.6296 & 0.6050 & 
  \textbf{0.4538} & 0.4504 & 0.4395 & 0.4142 & 0.4093 & 0.4280 & 
  0.5500 & \textbf{0.8101} & 0.7252 & 0.7044 & 0.6198 & 0.5975 \\ 
  
  mediamill & 
  0.5862 & \textbf{0.6791} & 0.6676 & 0.6244 & 0.6236 & 0.6066 & 
  \textbf{0.4240} & 0.3721 & 0.3557 & 0.3620 & 0.3607 & 0.3657 & 
  0.5862 & \textbf{0.6791} & 0.6676 & 0.6244 & 0.6236 & 0.6066 \\ 
  
  medical & 
  0.9389 & \textbf{0.9561} & \textbf{0.9561} & \textbf{0.9561} & \textbf{0.9561} & \textbf{0.9561} & 
  0.7488 & - & - & - & - & - & 
  0.9392 & \textbf{0.9560} & \textbf{0.9560} & \textbf{0.9560} & \textbf{0.9560} & \textbf{0.9560} \\ 
  
  tmc2007 & 
  \textbf{0.7727} & 0.7487 & 0.7584 & 0.7578 & 0.7577 & 0.7489 & 
  0.5482 & \textbf{0.6227} & 0.5592 & 0.5859 & 0.5911 & 0.5833 & 
  0.7658 & 0.7629 & \textbf{0.7696} & 0.7627 & 0.7581 & 0.7518 \\ 
  
  yeast & 
  0.6158 & 0.5563 & 0.5843 & 0.5918 & \textbf{0.6227} & 0.6166 & 
  \textbf{0.5760} & 0.5382 & 0.4899 & 0.4706 & 0.4908 & 0.5290 & 
  0.6239 & 0.5506 & 0.6005 & 0.6125 & 0.6478 & \textbf{0.6554} \\ 
  
  \midrule\Large{ML-kNN} \\
  cal500 & 
  \textbf{0.3417} & 0.1816 & 0.2045 & 0.2292 & 0.2703 & 0.3107 & 
  \textbf{0.3232} & 0.0812 & 0.0812 & 0.0812 & 0.0812 & 0.1292 & 
  \textbf{0.3250} & 0.0807 & 0.0816 & 0.0990 & 0.1271 & 0.1870 \\ 
  
  corel16k & 
  0.5759 & 0.6919 & 0.6919 & \textbf{0.6942} & 0.6348 & 0.6528 & 
  0.5545 & 0.5512 & 0.5512 & 0.5512 & 0.5512 & \textbf{0.5779} & 
  0.6327 & 0.6907 & \textbf{0.7035} & 0.6907 & 0.6841 & 0.6471 \\ 
  
  corel5k & 
  0.5826 & 0.5310 & \textbf{0.6214} & 0.6115 & 0.6038 & 0.5861 & 
  0.5860 & - & - & - & - & - & 
  0.5826 & 0.5310 & \textbf{0.6214} & 0.6115 & 0.6038 & 0.5861 \\ 
  
  delicious & 
  0.2816 & \textbf{0.3571} & 0.3401 & 0.2952 & 0.2809 & 0.2824 & 
  0.2756 & \textbf{0.4640} & 0.4338 & 0.3787 & 0.3609 & 0.3031 & 
  0.2816 & \textbf{0.3571} & 0.3401 & 0.2952 & 0.2809 & 0.2824 \\ 
  
  enron & 
  0.5851 & \textbf{0.6626} & 0.6376 & 0.5931 & 0.5603 & 0.5290 & 
  \textbf{0.5470} & 0.4816 & 0.4324 & 0.4324 & 0.4411 & 0.4772 & 
  0.5854 & \textbf{0.6954} & 0.6349 & 0.5810 & 0.5425 & 0.5231 \\ 
  
  mediamill & 
  \textbf{0.6492} & 0.4408 & 0.5273 & 0.5130 & 0.6004 & 0.6169 & 
  \textbf{0.5286} & 0.3514 & 0.3632 & 0.3585 & 0.3636 & 0.3627 & 
  \textbf{0.6492} & 0.4408 & 0.5273 & 0.5130 & 0.6004 & 0.6169 \\ 
  
  medical & 
  0.9098 & \textbf{0.9103} & \textbf{0.9103} & \textbf{0.9103} & \textbf{0.9103} & \textbf{0.9103} & 
  0.7702 & - & - & - & - & - & 
  \textbf{0.9196} & 0.9138 & 0.9138 & 0.9138 & 0.9138 & 0.9138 \\ 
  
  tmc2007 & 
  \textbf{0.7412} & 0.7064 & 0.7055 & 0.7029 & 0.7118 & 0.7244 & 
  0.6130 & 0.6202 & 0.5967 & 0.6208 & 0.6257 & \textbf{0.6261} & 
  0.7380 & \textbf{0.7627} & 0.7166 & 0.7032 & 0.7073 & 0.7195 \\ 
  
  yeast & 
  0.6536 & 0.5746 & 0.5945 & 0.5970 & 0.6566 & \textbf{0.6660} & 
  \textbf{0.6493} & 0.5761 & 0.4854 & 0.4567 & 0.4450 & 0.5543 & 
  \textbf{0.6995} & 0.5931 & 0.5560 & 0.5654 & 0.6471 & 0.6802 \\ 
   \bottomrule
\end{tabular}
\captionof{table}{F-Measure ($\uparrow$)}
\label{Table:FMeasure}
\end{landscape}

\begin{landscape}
\def\arraystretch{1.1}
\centering
\begin{tabular}{lrrrrrr|rrrrrr|rrrrrr}
  \hline
Classifier/ & \multicolumn{6}{c|}{ML-ROS} & \multicolumn{6}{c|}{MLeNN} & \multicolumn{6}{c}{MLSMOTE} \\ 
Dataset & 
          \multicolumn{1}{c}{Base} & \multicolumn{1}{c}{H$_{0.25}$} & \multicolumn{1}{c}{H$_{0.37}$} & \multicolumn{1}{c}{H$_{0.50}$} & \multicolumn{1}{c}{H$_{0.62}$} & \multicolumn{1}{c|}{H$_{0.75}$} & 
          \multicolumn{1}{c}{Base} & \multicolumn{1}{c}{H$_{0.25}$} & \multicolumn{1}{c}{H$_{0.37}$} & \multicolumn{1}{c}{H$_{0.50}$} & \multicolumn{1}{c}{H$_{0.62}$} & \multicolumn{1}{c|}{H$_{0.75}$} & 
          \multicolumn{1}{c}{Base} & \multicolumn{1}{c}{H$_{0.25}$} & \multicolumn{1}{c}{H$_{0.37}$} & \multicolumn{1}{c}{H$_{0.50}$} & \multicolumn{1}{c}{H$_{0.62}$} & \multicolumn{1}{c}{H$_{0.75}$} \\
  \toprule
  \Large{BR} \\ 
  cal500 & 
  0.6447 & \textbf{0.7104} & 0.7027 & 0.6960 & 0.6921 & 0.6933 & 
  0.6746 & 0.7509 & \textbf{0.7869} & 0.7496 & 0.7358 & 0.7133 & 
  0.7019 & \textbf{0.8147} & 0.8134 & 0.7979 & 0.7799 & 0.7565 \\ 
  
  chess & 
  0.7963 & 0.8169 & 0.8180 & \textbf{0.8192} & 0.8157 & 0.8138 & 
  \textbf{0.7138} & 0.3490 & 0.3540 & 0.3418 & 0.3431 & 0.3962 & 
  0.7963 & 0.8169 & 0.8180 & \textbf{0.8192} & 0.8157 & 0.8138 \\ 
  
  corel16k & 
  0.8091 & 0.8065 & 0.8076 & 0.8105 & \textbf{0.8115} & 0.8114 & 
  \textbf{0.6915} & 0.3273 & 0.2969 & 0.2880 & 0.3146 & 0.3877 & 
  0.8059 & 0.8057 & 0.8069 & 0.8098 & \textbf{0.8099} & 0.8092 \\ 
  
  corel5k & 
  0.8552 & 0.8588 & 0.8594 & \textbf{0.8601} & 0.8596 & 0.8593 & 
  \textbf{0.7005} & 0.4324 & 0.4183 & 0.4161 & 0.4163 & 0.4430 & 
  0.8552 & 0.8588 & 0.8594 & \textbf{0.8601} & 0.8596 & 0.8593 \\ 
  
  delicious & 
  0.8273 & 0.8441 & \textbf{0.8442} & \textbf{0.8442} & 0.8432 & 0.8401 & 
  \textbf{0.8231} & 0.6007 & 0.7513 & 0.8015 & 0.8198 & 0.8216 & 
  0.8273 & 0.8441 & \textbf{0.8442} & \textbf{0.8442} & 0.8432 & 0.8401 \\ 
  
  enron & 
  0.8231 & \textbf{0.8680} & 0.8675 & 0.8624 & 0.8525 & 0.8426 & 
  \textbf{0.7728} & 0.6863 & 0.6714 & 0.6807 & 0.7000 & 0.6409 & 
  0.8160 & 0.8657 & \textbf{0.8669} & 0.8568 & 0.8463 & 0.8382 \\ 
  
  mediamill & 
  0.8116 & \textbf{0.9190} & 0.9185 & 0.9157 & 0.9069 & 0.8845 & 
  \textbf{0.7076} & 0.3231 & 0.3219 & 0.3199 & 0.3222 & 0.3300 & 
  0.8116 & \textbf{0.9190} & 0.9185 & 0.9157 & 0.9069 & 0.8845 \\ 
  
  medical & 
  \textbf{0.9343} & 0.9306 & 0.9306 & 0.9306 & 0.9306 & 0.9306 & 
  \textbf{0.6780} & 0.3865 & 0.3865 & 0.3865 & 0.3865 & 0.3865 & 
  \textbf{0.9283} & 0.9261 & 0.9261 & 0.9261 & 0.9261 & 0.9261 \\ 
  
  tmc2007 & 
  0.8893 & 0.8698 & 0.8727 & 0.8984 & \textbf{0.9033} & 0.9023 & 
  \textbf{0.7974} & 0.3393 & 0.6961 & 0.6505 & 0.6325 & 0.6332 & 
  \textbf{0.8783} & 0.8427 & 0.8445 & 0.8557 & 0.8579 & 0.8699 \\ 
  
  yeast & 
  0.6909 & \textbf{0.7341} & 0.7227 & 0.7161 & 0.7148 & 0.7163 & 
  \textbf{0.6618} & 0.5850 & 0.5264 & 0.5077 & 0.5091 & 0.6395 & 
  0.6863 & \textbf{0.7854} & 0.7731 & 0.7622 & 0.7519 & 0.7459 \\ 
  
  \midrule\Large{LP} \\
  cal500 & 
  \textbf{0.6077} & 0.5474 & 0.5505 & 0.5611 & 0.5621 & 0.5817 & 
  0.4246 & \textbf{0.4758} & 0.4648 & 0.4756 & 0.4578 & 0.4379 & 
  0.4342 & \textbf{0.4632} & 0.4569 & 0.4451 & 0.4415 & 0.4550 \\ 
  
  chess & 
  0.4758 & 0.4974 & 0.4916 & 0.4937 & \textbf{0.5015} & 0.4929 & 
  0.4747 & 0.5013 & 0.5008 & 0.5023 & 0.5009 & \textbf{0.5055} & 
  0.4758 & 0.4974 & 0.4916 & 0.4937 & \textbf{0.5015} & 0.4929 \\ 
  
  corel16k & 
  0.4863 & 0.4972 & \textbf{0.4980} & 0.4969 & 0.4964 & 0.4943 & 
  0.4798 & 0.5025 & 0.5017 & \textbf{0.5026} & 0.5011 & 0.5021 & 
  0.4757 & \textbf{0.5001} & 0.4983 & 0.4967 & 0.4947 & 0.4931 \\ 
  
  corel5k & 
  0.4717 & 0.5004 & 0.5010 & \textbf{0.5019} & 0.5003 & 0.4961 & 
  0.4890 & 0.5009 & 0.5007 & 0.5018 & 0.5005 & \textbf{0.5021} & 
  0.4717 & 0.5004 & 0.5010 & \textbf{0.5019} & 0.5003 & 0.4961 \\ 
  
  delicious & 
  0.4238 & \textbf{0.4826} & 0.4740 & 0.4643 & 0.4560 & 0.4473 & 
  0.4275 & \textbf{0.5014} & 0.4965 & 0.4906 & 0.4831 & 0.4679 & 
  0.4238 & \textbf{0.4826} & 0.4740 & 0.4643 & 0.4560 & 0.4473 \\ 
  
  enron & 
  0.5437 & 0.5517 & 0.5531 & 0.5630 & \textbf{0.5680} & 0.5601 & 
  0.5007 & 0.5079 & 0.5014 & 0.4740 & 0.4959 & \textbf{0.5116} & 
  0.5364 & 0.5507 & 0.5544 & 0.5642 & 0.5627 & \textbf{0.5655} \\ 
  
  mediamill & 
  \textbf{0.6610} & 0.5421 & 0.5985 & 0.6106 & 0.6360 & 0.6477 & 
  \textbf{0.5489} & 0.5047 & 0.5031 & 0.5059 & 0.5062 & 0.5028 & 
  \textbf{0.6610} & 0.5421 & 0.5985 & 0.6106 & 0.6360 & 0.6477 \\ 
  
  medical & 
  \textbf{0.8901} & 0.8567 & 0.8567 & 0.8567 & 0.8567 & 0.8567 & 
  \textbf{0.6059} & 0.4948 & 0.4948 & 0.4948 & 0.4948 & 0.4948 & 
  \textbf{0.8872} & 0.8486 & 0.8486 & 0.8486 & 0.8486 & 0.8486 \\ 
  
  tmc2007 & 
  \textbf{0.8066} & 0.5932 & 0.6287 & 0.6624 & 0.6969 & 0.7228 & 
  \textbf{0.6842} & 0.5158 & 0.5754 & 0.5364 & 0.5384 & 0.5326 & 
  \textbf{0.7881} & 0.5703 & 0.6120 & 0.6474 & 0.6840 & 0.7191 \\ 
  
  yeast & 
  \textbf{0.6684} & 0.5626 & 0.5944 & 0.6034 & 0.6482 & 0.6570 & 
  \textbf{0.6354} & 0.5654 & 0.5057 & 0.4944 & 0.4956 & 0.5677 & 
  \textbf{0.6717} & 0.5241 & 0.5766 & 0.5985 & 0.6451 & 0.6686 \\ 
  
  \midrule\Large{ML-kNN} \\
  cal500 & 
  0.7723 & 0.7735 & \textbf{0.7726} & 0.7714 & 0.7708 & 0.7708 & 
  \textbf{0.8081} & 0.7448 & 0.7900 & 0.7825 & 0.8014 & 0.8048 & 
  \textbf{0.8136} & 0.8109 & 0.8116 & 0.8117 & 0.8124 & 0.8132 \\ 
  
  chess & 
  \textbf{0.8349} & 0.8314 & 0.8321 & 0.8328 & 0.8335 & 0.8339 & 
  \textbf{0.7355} & 0.3472 & 0.3516 & 0.3376 & 0.3433 & 0.3979 & 
  \textbf{0.8349} & 0.8314 & 0.8321 & 0.8328 & 0.8335 & 0.8339 \\ 
  
  corel16k & 
  \textbf{0.8236} & 0.8203 & 0.8208 & 0.8214 & 0.8221 & 0.8225 &
  \textbf{0.7003} & 0.3298 & 0.2998 & 0.2941 & 0.3232 & 0.3974 & 
  \textbf{0.8250} & 0.8213 & 0.8218 & 0.8224 & 0.8234 & 0.8238 \\ 
  
  corel5k & 
  0.8675 & 0.8670 & 0.8675 & 0.8675 & 0.8677 & \textbf{0.8676} & 
  \textbf{0.7083} & 0.4319 & 0.4176 & 0.4152 & 0.4153 & 0.4455 & 
  0.8675 & 0.8670 & 0.8675 & 0.8675 & 0.8677 & \textbf{0.8676} \\ 
  
  delicious & 
  \textbf{0.8788} & 0.8753 & 0.8759 & 0.8765 & 0.8772 & 0.8779 & 
  \textbf{0.8759} & 0.6274 & 0.7750 & 0.8302 & 0.8535 & 0.8665 & 
  \textbf{0.8788} & 0.8753 & 0.8759 & 0.8765 & 0.8772 & 0.8779 \\ 
  
  enron & 
  \textbf{0.9007} & 0.8984 & 0.8991 & 0.8995 & 0.8999 & 0.8997 & 
  \textbf{0.8421} & 0.7147 & 0.7402 & 0.7564 & 0.7830 & 0.7136 & 
  \textbf{0.9016} & 0.8990 & 0.8999 & 0.9003 & 0.9011 & 0.9011 \\ 
  
  mediamill & 
  \textbf{0.9552} & 0.9533 & 0.9537 & 0.9539 & 0.9544 & 0.9545 & 
  \textbf{0.8746} & 0.3198 & 0.3189 & 0.3157 & 0.3157 & 0.3191 & 
  \textbf{0.9552} & 0.9533 & 0.9537 & 0.9539 & 0.9544 & 0.9545 \\ 
  
  medical & 
  \textbf{0.9515} & 0.9488 & 0.9488 & 0.9488 & 0.9488 & 0.9488 & 
  \textbf{0.6787} & 0.3404 & 0.3404 & 0.3404 & 0.3404 & 0.3404 & 
  \textbf{0.9553} & 0.9511 & 0.9511 & 0.9511 & 0.9511 & 0.9511 \\ 
  
  tmc2007 & 
  \textbf{0.9368} & 0.9306 & 0.9308 & 0.9315 & 0.9311 & 0.9312 & 
  \textbf{0.8483} & 0.3323 & 0.6845 & 0.5743 & 0.5706 & 0.5902 & 
  \textbf{0.9419} & 0.9337 & 0.9343 & 0.9359 & 0.9362 & 0.9373 \\ 
  
  yeast & 
  \textbf{0.8081} & 0.7997 & 0.7985 & 0.7983 & 0.8029 & 0.8030 & 
  \textbf{0.8052} & 0.6380 & 0.5262 & 0.4814 & 0.4793 & 0.7161 & 
  \textbf{0.8466} & 0.8192 & 0.8249 & 0.8260 & 0.8346 & 0.8404 \\ 
   \bottomrule
\end{tabular}
\captionof{table}{AUC ($\uparrow$)}
\label{Table:AUC}
\end{landscape}

\begin{itemize}
    \item \textbf{Ranking Loss:} By examining the Table \ref{Table:RankingLoss} is easy to extract some clear conclusions. The hybrid version achieves most of the best results when paired with the BR classifier. As many as 8 or 9 out of 10 cases have improved the base result. On the contrary, the hybridization does not benefit the MLeNN resampling method, nor the behavior of the ML-kNN classifier. The situation with the LP classifier is in between, with half of the hybrid configurations improving and the other half worsening.

    \item \textbf{Precision:} From Table \ref{Table:Precision} two main facts can be drawn. That for nearly all cases hybrid configurations achieve most of the best results is the first one. In some cases, such as BR and ML-kNN, almost all best values correspond to the hybrid versions of ML-ROS and MLSMOTE. The second, that the hybridization is not able to improve the behavior of MLeNN. 

    \item \textbf{F-measure:} The F-measure values collected in Table \ref{Table:FMeasure} show a scenario quite similar for all configurations. In most of them, hybrid versions of the resampling methods achieve more best results, including MLeNN.

    \item \textbf{AUC:} As can be seen in Table \ref{Table:AUC}, the distribution of best values is quite similar to that of the Ranking Loss (Table \ref{Table:RankingLoss}) metric. The proposed hybridization works best when applied to ML-ROS and MLSMOTE and combined with BR and LP. On the other side, it does not benefit the work of MLeNN, and it does not seem to mix well with the ML-kNN classifier.
\end{itemize}

From a global perspective, Hamming Loss, Precision and F-Measure provide a close evaluation of the results, with the hybrid resampling achieving between 60\%-70\% of best values. By contrast, the assessment offered by Ranking Loss and AUC indicates that the hybridization improves results in slightly less than 50\% of the cases. Nevertheless, aside from this overall view what it is interesting to scrutinize is how these best values are distributed according to resampling and classification methods.

Focusing in the resampling methods, the further consequences can be drawn:

\begin{itemize}
    \item \textbf{ML-ROS and MLSMOTE:} As have been mentioned, both algorithms are, in general, able to benefit from the hibridization, as can be stated from the tables of results observation. Working over decoupled data samples, ML-ROS and MLSMOTE can gather instances associated to minority labels that do not include also majority ones, being capable of producing more samples exclusively linked to minority labels.
    
    \item \textbf{MLeNN:} As we have highlighted, MLeNN is not benefiting of the label decoupling process. The reason to this behavior can be deducted by inspecting the inner workings of MLeNN. This algorithm compares the labelset of the selected instance with those of their nearest neighbors, removing the pattern if there is a significant difference with more than half of these neighbors. Since the decoupling process locates two instances with disjoint labelsets into exactly the same position (they share the feature set), it is effectively increasing the likelihood of removing the majority sample just produced by the split. It seems that the loss of information generated by this fact is enough to deteriorate the classifier performance. 

\end{itemize}

Regarding the behavior of MLC algorithms, the hybridization produces the following effects:

\begin{itemize}
    \item \textbf{BR and LP:} The results produced by BR and LP classifiers improve the ones of the base resampling in a large portion of the studied cases. In general the BR approach seems to be the most benefited, with about an 80\% of hybrid configurations achieving best results after taking into account ML-ROS and MLSMOTE only. Considering that BR trains an independent binary classifier for each label, by splitting minority and majority labels in separate instances the performance of these individual classifiers improve. The influence on the LP classifier comes from the fact that by decoupling imbalanced labels the global amount of label combinations is also reduced. As a result, the base multiclass classifier has to deal with less classes.
    
    \item \textbf{ML-kNN:} If the interest is in getting a good Precision and F-Measure, the results of hybrid versions with ML-kNN are quite remarkable. However, the results with AUC, Ranking Loss, and partially also with Hamming Loss allow to infer that this is not a good mixture, independently of the resampling being applied. This behavior could be due to similar reasons to that explained above for MLeNN. ML-kNN starts by computing a priori probabilities for each label, whose result should not be affected by the decoupling process. However, each time a new sample is to be classified the algorithm has to find their nearest neighbors. Is in this operation when the problem might arise, since there will be two instances located at the same distance but with a disjoint set of labels. As a result, the computing of the a posteriori probabilities of ML-kNN will be affected.
\end{itemize}

Lastly, the results are briefly analyzed regarding the goodness of the proposed hybridization for the selected MLDs. Some consequences can be also deduced.
\begin{itemize}
    \item Six out of the ten used MLDs, cal500, chess, corel16k, corel5k, delicious, and enron, usually improve the results after the proposed hybridization has been applied. These MLDs have the proper traits to benefit from the label decoupling, since they have a significant level of label concurrence as their \textit{SCUMBLE} values denote, and also from the resampling, as all of them are imbalanced datasets.
    
    \item The results for the datasets medical, tmc2007 and yeast are usually obtained by the original preprocessing, instead of with one of the hybrid configurations. By examining the traits of these datasets (see Table \ref{MLDs}), that they have the lowest \textit{SCUMBLE} values can be stated. In fact, these values are below or marginally above the threshold indicated in \cite{Charte:REMEDIAL} as recommendation to apply REMEDIAL. A low \textit{SCUMBLE} value denotes that the MLD does not suffer from imbalanced labels concurrence, so the REMEDIAL algorithm would have little impact if any. As a consequence, the differences in classification performance tend to be quite slight, mostly due to the non-deterministic behavior of the classifier.
    
    \item Although less frequently than the three prior cases, the mediamill MLD also presents the aforementioned demeanor.  However, the case of mediamill is not comparable since it has a high \textit{SCUMBLE} value. Notwithstanding, this MLD presents the highest imbalance level as can be seen in the MeanIR column of Table \ref{MLDs}. So, it is an MLD inherently hard to learn from, and sometimes the decoupling does not positively contribute the training of the classifiers.
\end{itemize}

As regards to the threshold level that determines which data samples are processed by REMEDIAL, in general values in the $[0.25, 0.50]$ interval are producing better results than those above 0.50. This seems coherent, since the higher is the threshold the fewer samples will be decoupled. However, taking too many instances can be also detrimental to the posterior resampling. The best cut value will be influence by the MLD traits, as well as the chosen classifier and resampling algorithms. So, it should be adjusted taking into account all these variables, maybe through an internal cross validation step aimed to optimize this parameter.

\section{Conclusions}\label{Conclusions}
In this work we have proposed three hybrid preprocessing methods aimed to tackle imbalanced multilabel learning. The conducted experimentation has demonstrated its effectiveness. In addition, from the study of interactions between REMEDIAL, resampling methods and classifiers, a set of clear guidelines on when the use of the hybridization is beneficial can be extracted:
\begin{itemize}
    \item Combining imbalanced labels decoupling with data resampling is positive as long as it is applied to MLDs having a high label concurrence problem. Otherwise, the effect of the hybrid preprocessings can be negligible or even induce a worsening of the results. This is a conclusion already argued in \cite{Charte:REMEDIAL}.
    
    \item The proposed hybridization improves the efficiency of oversampling algorithms, such as ML-ROS and MLSMOTE, as they will be able to produce new instances that include only the minority labels. On the contrary, it should not be used with methods such as MLeNN, based on locating nearest neighbors with high differences in the labelsets to remove them.
    
    \item The MLDs preprocessed with hybrid resampling methods improve the training of BR and LP classifiers, and it is reasonable to assume that this improvement would be also applicable to BR-based and LP-based methods. On the other side, classifiers such as ML-kNN, which are based on nearest neighbor information, could degrade their performance. This is due to the fact that the splitted instances are located exactly in the same position.
    
    \item Depending on the selected evaluation metric, the obtained view about how classifiers perform can drastically change. The hybridization should be chosen if boosting Precision or F-Measure is the goal, no matter which resample or classifier it is tied to. If the objective is to maximize AUC or minimize Ranking Loss, the decoupling should only be combined with oversampling and BR/LP classifiers.
\end{itemize}

Summarizing, the label decoupling plus data resampling combination has a positive impact in classification results as long as certain conditions are met. Firstly, the decoupling should only be applied to MLDs having a high level of imbalanced labels concurrence. Second, that the proposed methodology is able to improve classification performance mostly when paired with oversampling techniques and BR and LP algorithms.

The main obstacle to achieve a more general gain from the proposed hybridization comes from the fact that decoupled instances, although they have separate minority and majority labels and this helps some resampling methods, are located in the same position (their set of input features does not change). A pair of aspects could worth further study:
\begin{itemize}
    \item A potential solution for the mentioned problems with MLeNN and ML-kNN would be enhancing the REMEDIAL algorithm, thus that after the splitting the resulting instances are relocated according to their new labelsets.
    
    \item In addition, the threshold to decide when a data sample should be decoupled or not could be automatically adjusted, for instance through cross validation techniques.
\end{itemize} 

\textbf{Acknowledgments}: This work is partially supported by the Spanish Ministry of Science and Technology under projects TIN2014-57251-P and TIN2012-33856, and the Andalusian regional project P11-TIC-7765.

%\bibliographystyle{elsarticle-num}
%\bibliography{fcharte}

\section*{References}
\begin{thebibliography}{10}
\expandafter\ifx\csname url\endcsname\relax
  \def\url#1{\texttt{#1}}\fi
\expandafter\ifx\csname urlprefix\endcsname\relax\def\urlprefix{URL }\fi
\expandafter\ifx\csname href\endcsname\relax
  \def\href#1#2{#2} \def\path#1{#1}\fi

\bibitem{Aggarwal:2014}
C.~C. Aggarwal, Data classification: algorithms and applications, CRC Press,
  2014.

\bibitem{fayyad1996data}
U.~Fayyad, G.~Piatetsky-Shapiro, P.~Smyth, From data mining to knowledge
  discovery in databases, AI magazine 17~(3) (1996) 37.

\bibitem{Tsoumakas3}
G.~Tsoumakas, I.~Katakis, I.~Vlahavas, {Mining Multi-label Data}, in:
  O.~Maimon, L.~Rokach (Eds.), Data Mining and Knowledge Discovery Handbook,
  Springer US, Boston, MA, 2010, Ch.~34, pp. 667--685.
\newblock \href {http://dx.doi.org/10.1007/978-0-387-09823-4\_34}
  {\path{doi:10.1007/978-0-387-09823-4\_34}}.

\bibitem{Charte:SB-MLC}
F.~Herrera, F.~Charte, A.~J. Rivera, M.~J. del Jesus, {Multilabel
  Classification. Problem analysis, metrics and techniques}, Springer, 2016.
\newblock \href {http://dx.doi.org/10.1007/978-3-319-41111-8}
  {\path{doi:10.1007/978-3-319-41111-8}}.

\bibitem{ReviewVentura}
E.~Gibaja, S.~Ventura, Multi-label learning: a review of the state of the art
  and ongoing research, Wiley Interdisciplinary Reviews: Data Mining and
  Knowledge Discovery 4~(6) (2014) 411--444.
\newblock \href {http://dx.doi.org/10.1002/widm.1139}
  {\path{doi:10.1002/widm.1139}}.

\bibitem{TutorialVentura}
E.~Gibaja, S.~Ventura, A tutorial on multilabel learning, ACM Computing Surveys
  47~(3) (2015) 52:1--52:38.
\newblock \href {http://dx.doi.org/10.1145/2716262}
  {\path{doi:10.1145/2716262}}.

\bibitem{Glinka2016}
K.~Glinka, A.~Wosiak, D.~Zakrzewska, {Improving Children Diagnostics by
  Efficient Multi-label Classification Method}, in: Advances in Intelligent
  Systems and Computing, Vol. 471, Springer International Publishing, 2016, pp.
  253--266.
\newblock \href {http://dx.doi.org/10.1007/978-3-319-39796-2\_21}
  {\path{doi:10.1007/978-3-319-39796-2\_21}}.

\bibitem{QUINTA}
F.~Charte, A.~J. Rivera, M.~J. del Jesus, F.~Herrera, {QUINTA: A question
  tagging assistant to improve the answering ratio in electronic forums}, in:
  EUROCON 2015 - International Conference on Computer as a Tool (EUROCON),
  IEEE, 2015, pp. 1--6.
\newblock \href {http://dx.doi.org/10.1109/EUROCON.2015.7313677}
  {\path{doi:10.1109/EUROCON.2015.7313677}}.

\bibitem{Wei2013}
Z.~Wei, H.~Wang, R.~Zhao, {Semi-supervised multi-label image classification
  based on nearest neighbor editing}, Neurocomputing 119 (2013) 462--468.
\newblock \href {http://dx.doi.org/10.1016/j.neucom.2013.03.011}
  {\path{doi:10.1016/j.neucom.2013.03.011}}.

\bibitem{Che2016}
Y.~Che, Y.~Ju, P.~Xuan, R.~Long, F.~Xing, {Identification of multi-functional
  Enzyme with multi-label classifier}, PLoS ONE 11~(4) (2016) e0153503.
\newblock \href {http://dx.doi.org/10.1371/journal.pone.0153503}
  {\path{doi:10.1371/journal.pone.0153503}}.

\bibitem{Alberto:2013}
A.~Fern\'andez, V.~L\'opez, M.~Galar, M.~J. del Jesus, F.~Herrera, Analysing
  the classification of imbalanced data-sets with multiple classes:
  Binarization techniques and ad-hoc approaches, Knowl. Based Systems 42 (2013)
  97 -- 110.
\newblock \href {http://dx.doi.org/10.1016/j.knosys.2013.01.018}
  {\path{doi:10.1016/j.knosys.2013.01.018}}.

\bibitem{He2009}
H.~He, E.~A. Garcia, {Learning from imbalanced data}, IEEE Transactions on
  Knowledge and Data Engineering 21~(9) (2009) 1263--1284.
\newblock \href {http://dx.doi.org/10.1109/TKDE.2008.239}
  {\path{doi:10.1109/TKDE.2008.239}}.

\bibitem{Lopez:2013}
V.~L\'opez, A.~Fern\'andez, S.~Garc\'ia, V.~Palade, F.~Herrera, An insight into
  classification with imbalanced data: Empirical results and current trends on
  using data intrinsic characteristics, Inf. Sciences 250 (2013) 113 -- 141.
\newblock \href {http://dx.doi.org/10.1016/j.ins.2013.07.007}
  {\path{doi:10.1016/j.ins.2013.07.007}}.

\bibitem{Prati2014}
R.~C. Prati, G.~E. A. P.~A. Batista, D.~F. Silva, {Class imbalance revisited: a
  new experimental setup to assess the performance of treatment methods},
  Knowledge and Information Systems 45~(1) (2014) 247--270.
\newblock \href {http://dx.doi.org/10.1007/s10115-014-0794-3}
  {\path{doi:10.1007/s10115-014-0794-3}}.

\bibitem{Charte:Neucom13}
F.~Charte, A.~J. Rivera, M.~J. del Jesus, F.~Herrera, Addressing imbalance in
  multilabel classification: Measures and random resampling algorithms,
  Neurocomputing 163~(0) (2015) 3--16.
\newblock \href {http://dx.doi.org/10.1016/j.neucom.2014.08.091}
  {\path{doi:10.1016/j.neucom.2014.08.091}}.

\bibitem{JSSanchez:13}
V.~Garc\'ia, J.~S\'anchez, R.~Mollineda, On the effectiveness of preprocessing
  methods when dealing with different levels of class imbalance, Knowl. Based
  Systems 25~(1) (2012) 13 -- 21.
\newblock \href
  {http://dx.doi.org/http://dx.doi.org/10.1016/j.knosys.2011.06.013}
  {\path{doi:http://dx.doi.org/10.1016/j.knosys.2011.06.013}}.

\bibitem{Giraldo:2013}
A.~F. Giraldo-Forero, J.~A. Jaramillo-Garz\'on, J.~F. Ruiz-Mu\~noz, C.~G.
  Castellanos-Dom\'inguez, Managing imbalanced data sets in multi-label
  problems: A case study with the {SMOTE} algorithm, in: Progress in Pattern
  Recognit., Image Analysis, Computer Vision, and Applications, Vol. 8258 of
  LNCS, Springer, 2013, pp. 334--342.
\newblock \href {http://dx.doi.org/10.1007/978-3-642-41822-8\_42}
  {\path{doi:10.1007/978-3-642-41822-8\_42}}.

\bibitem{Charte:IDEAL14}
F.~Charte, A.~Rivera, M.~del Jesus, F.~Herrera, Mlenn: A first approach to
  heuristic multilabel undersampling, in: Proc. 15th Int. Conf. Intelligent
  Data Engineering and Automated Learning, Salamanca, Spain, IDEAL'14, Vol.
  8669 of LNCS, 2014, pp. 1--9.
\newblock \href {http://dx.doi.org/10.1007/978-3-319-10840-7\_1}
  {\path{doi:10.1007/978-3-319-10840-7\_1}}.

\bibitem{Charte:MLSMOTE}
F.~Charte, A.~J. Rivera, M.~J. del Jesus, F.~Herrera, {MLSMOTE: Approaching
  imbalanced multilabel learning through synthetic instance generation},
  Knowledge-Based Systems 89 (2015) 385--397.
\newblock \href {http://dx.doi.org/10.1016/j.knosys.2015.07.019}
  {\path{doi:10.1016/j.knosys.2015.07.019}}.

\bibitem{Charte:HAIS14}
F.~Charte, A.~Rivera, M.~J. del Jesus, F.~Herrera, Concurrence among imbalanced
  labels and its influence on multilabel resampling algorithms, in: Hybrid
  Artificial Intelligence Systems, Vol. 8480 of LNCS, Springer International
  Publishing, 2014, pp. 110--121.
\newblock \href {http://dx.doi.org/10.1007/978-3-319-07617-1\_10}
  {\path{doi:10.1007/978-3-319-07617-1\_10}}.

\bibitem{Charte:REMEDIAL}
F.~Charte, A.~Rivera, M.~J. del Jesus, F.~Herrera, {Resampling Multilabel
  Datasets by Decoupling Highly Imbalanced Labels}, in: Hybrid Artificial
  Intelligent Systems, Vol. 9121 of Lecture Notes in Computer Science, Springer
  International Publishing, 2015, pp. 489--501.
\newblock \href {http://dx.doi.org/10.1007/978-3-319-19644-2\_41}
  {\path{doi:10.1007/978-3-319-19644-2\_41}}.

\bibitem{Elghazel2016}
H.~Elghazel, A.~Aussem, O.~Gharroudi, W.~Saadaoui, {Ensemble multi-label text
  categorization based on rotation forest and latent semantic indexing}, Expert
  Systems with Applications 57 (2016) 1--11.
\newblock \href {http://dx.doi.org/10.1016/j.eswa.2016.03.041}
  {\path{doi:10.1016/j.eswa.2016.03.041}}.

\bibitem{Jing2016}
X.-y. Jing, F.~Wu, Z.~Li, R.~Hu, S.~Member, D.~Zhang, {Multi-Label Dictionary
  Learning for Image Annotation}, IEEE Transactions on Image Processing 25~(6)
  (2016) 2712--2725.
\newblock \href {http://dx.doi.org/10.1109/TIP.2016.2549459}
  {\path{doi:10.1109/TIP.2016.2549459}}.

\bibitem{Godbole}
S.~Godbole, S.~Sarawagi, {Discriminative Methods for Multi-Labeled
  Classification}, in: Advances in Knowl. Discovery and Data Mining, Vol. 3056,
  2004, pp. 22--30.
\newblock \href {http://dx.doi.org/10.1007/978-3-540-24775-3\_5}
  {\path{doi:10.1007/978-3-540-24775-3\_5}}.

\bibitem{Boutell}
M.~Boutell, J.~Luo, X.~Shen, C.~Brown, {Learning multi-label scene
  classification}, Pattern Recognit. 37~(9) (2004) 1757--1771.
\newblock \href {http://dx.doi.org/10.1016/j.patcog.2004.03.009}
  {\path{doi:10.1016/j.patcog.2004.03.009}}.

\bibitem{mencia2010efficient}
E.~{Loza Menc{\'{i}}a}, S.~H. Park, J.~F{\"{u}}rnkranz, {Efficient voting
  prediction for pairwise multilabel classification}, Neurocomputing 73~(7-9)
  (2010) 1164--1176.
\newblock \href {http://dx.doi.org/10.1016/j.neucom.2009.11.024}
  {\path{doi:10.1016/j.neucom.2009.11.024}}.

\bibitem{Clare}
A.~Clare, R.~D. King, Knowledge discovery in multi-label phenotype data, in:
  Proc. 5th European Conf. Principles on Data Mining and Knowl. Discovery,
  Freiburg, Germany, PKDD'01, Vol. 2168, 2001, pp. 42--53.
\newblock \href {http://dx.doi.org/10.1007/3-540-44794-6\_4}
  {\path{doi:10.1007/3-540-44794-6\_4}}.

\bibitem{Zhang1}
M.~Zhang, Z.~Zhou, {ML-KNN: A lazy learning approach to multi-label learning},
  Pattern Recognit. 40~(7) (2007) 2038--2048.
\newblock \href {http://dx.doi.org/10.1016/j.patcog.2006.12.019}
  {\path{doi:10.1016/j.patcog.2006.12.019}}.

\bibitem{Elisseeff1}
A.~Elisseeff, J.~Weston, {A Kernel Method for Multi-Labelled Classification},
  in: Advances in Neural Information Processing Systems 14, Vol.~14, MIT Press,
  2001, pp. 681--687.

\bibitem{Zhou:MIML:2009}
M.-L. Zhang, Z.-J. Wang, {MIMLRBF: RBF} neural networks for multi-instance
  multi-label learning, Neurocomputing 72~(16–-18) (2009) 3951 -- 3956.
\newblock \href {http://dx.doi.org/10.1016/j.neucom.2009.07.008}
  {\path{doi:10.1016/j.neucom.2009.07.008}}.

\bibitem{Tahir:2012}
M.~A. Tahir, J.~Kittler, F.~Yan, Inverse random under sampling for class
  imbalance problem and its application to multi-label classification, Pattern
  Recognit. 45~(10) (2012) 3738--3750.
\newblock \href {http://dx.doi.org/10.1016/j.patcog.2012.03.014}
  {\path{doi:10.1016/j.patcog.2012.03.014}}.

\bibitem{Tepvorachai:2008}
G.~Tepvorachai, C.~Papachristou, Multi-label imbalanced data enrichment process
  in neural net classifier training, in: IEEE Int. Joint Conf. on Neural
  Networks, 2008. IJCNN, 2008, pp. 1301--1307.
\newblock \href {http://dx.doi.org/10.1109/IJCNN.2008.4633966}
  {\path{doi:10.1109/IJCNN.2008.4633966}}.

\bibitem{He:2012}
J.~He, H.~Gu, W.~Liu, Imbalanced multi-modal multi-label learning for
  subcellular localization prediction of human proteins with both single and
  multiple sites, PloS one 7~(6) (2012) 7155.
\newblock \href {http://dx.doi.org/10.1371/journal.pone.0037155}
  {\path{doi:10.1371/journal.pone.0037155}}.

\bibitem{LISHI:2013}
C.~Li, G.~Shi, Improvement of learning algorithm for the multi-instance
  multi-label rbf neural networks trained with imbalanced samples, J. Inf. Sci.
  Eng. 29~(4) (2013) 765--776.

\bibitem{Tahir:2012:2}
M.~A. Tahir, J.~Kittler, A.~Bouridane, Multilabel classification using
  heterogeneous ensemble of multi-label classifiers, Pattern Recognit. Lett.
  33~(5) (2012) 513--523.
\newblock \href {http://dx.doi.org/10.1016/j.patrec.2011.10.019}
  {\path{doi:10.1016/j.patrec.2011.10.019}}.

\bibitem{Charte:mldr}
F.~Charte, D.~Charte, Working with multilabel datasets in {R}: The mldr
  package, The R Journal 7~(2) (2015) 149--162.

\bibitem{Charte:RUMDR}
F.~Charte, D.~Charte, A.~J. Rivera, M.~J. del Jesus, F.~Herrera, {R Ultimate
  Multilabel Dataset Repository}, in: Proc. 11th International Conference on
  Hybrid Artificial Intelligent Systems, HAIS'16, Vol. 9648, Springer, 2016,
  pp. 487--499.
\newblock \href {http://dx.doi.org/10.1007/978-3-319-32034-2\_41}
  {\path{doi:10.1007/978-3-319-32034-2\_41}}.

\bibitem{Charte:HAIS16}
F.~Charte, A.~J. Rivera, M.~J. del Jesus, F.~Herrera, {On the Impact of Dataset
  Complexity and Sampling Strategy in Multilabel Classifiers Performance}, in:
  Proc. 11th International Conference on Hybrid Artificial Intelligent Systems,
  HAIS'16, Vol. 9648, Springer, 2016, pp. 500--511.
\newblock \href {http://dx.doi.org/10.1007/978-3-319-32034-2\_42}
  {\path{doi:10.1007/978-3-319-32034-2\_42}}.

\bibitem{Read}
J.~Read, B.~Pfahringer, G.~Holmes, E.~Frank, {Classifier chains for multi-label
  classification}, Mach. Learn. 85 (2011) 333--359.
\newblock \href {http://dx.doi.org/10.1007/s10994-011-5256-5}
  {\path{doi:10.1007/s10994-011-5256-5}}.

\bibitem{RPC}
E.~H{\"u}llermeier, J.~F{\"u}rnkranz, W.~Cheng, K.~Brinker, Label ranking by
  learning pairwise preferences, Artificial Intelligence 172~(16) (2008)
  1897--1916.
\newblock \href {http://dx.doi.org/10.1016/j.artint.2008.08.002}
  {\path{doi:10.1016/j.artint.2008.08.002}}.

\bibitem{CLR}
J.~F\"{u}rnkranz, E.~H\"{u}llermeier, E.~Loza~Menc\'{\i}a, K.~Brinker,
  Multilabel classification via calibrated label ranking, Mach. Learn. 73
  (2008) 133--153.
\newblock \href {http://dx.doi.org/10.1007/s10994-008-5064-8}
  {\path{doi:10.1007/s10994-008-5064-8}}.

\bibitem{Read:2008}
J.~Read, A pruned problem transformation method for multi-label classification,
  in: Proc. 2008 New Zealand Computer Science Research Student Conference
  (NZCSRS 2008), 2008, pp. 143--150.

\bibitem{Read:2008:2}
J.~Read, B.~Pfahringer, G.~Holmes, Multi-label classification using ensembles
  of pruned sets, in: Proc. 8th IEEE Int. Conf. on Data Mining, Pisa, Italy,
  ICDM'08, 2008, pp. 995--1000.

\bibitem{HOMER}
G.~Tsoumakas, I.~Katakis, I.~Vlahavas, {Effective and Efficient Multilabel
  Classification in Domains with Large Number of Labels}, in: Proc. ECML/PKDD
  Workshop on Mining Multidimensional Data, Antwerp, Belgium, MMD'08, 2008, pp.
  30--44.

\bibitem{Tsoumakas4}
G.~Tsoumakas, I.~Vlahavas, Random k-labelsets: An ensemble method for
  multilabel classification, in: Proc. 18th European Conf. on Machine Learning,
  Warsaw, Poland, ECML'07, Vol. 4701, 2007, pp. 406--417.
\newblock \href {http://dx.doi.org/10.1007/978-3-540-74958-5\_38}
  {\path{doi:10.1007/978-3-540-74958-5\_38}}.

\bibitem{Cheng}
W.~Cheng, E.~H\"{u}llermeier, {Combining instance-based learning and logistic
  regression for multilabel classification}, Mach. Learn. 76~(2-3) (2009)
  211--225.
\newblock \href {http://dx.doi.org/10.1007/s10994-009-5127-5}
  {\path{doi:10.1007/s10994-009-5127-5}}.

\bibitem{MULAN}
G.~Tsoumakas, E.~S. Xioufis, J.~Vilcek, I.~Vlahavas, {MULAN: A Java Library for
  Multi-Label Learning}, J. Mach. Learn. Res. 12 (2011) 2411--2414.
\end{thebibliography}
\end{document}